\documentclass[a4paper,fleqn]{cas-dc}

\usepackage[numbers]{natbib}
\usepackage{svg}
\usepackage{graphicx}
\usepackage{subcaption}
\usepackage{float}
\usepackage{hyperref}
\usepackage{doi}
\usepackage{pifont}

\def\tsc#1{\csdef{#1}{\textsc{\lowercase{#1}}\xspace}}
\tsc{WGM}
\tsc{QE}
\begin{document}
\let\WriteBookmarks\relax
\def\floatpagepagefraction{1}
\def\textpagefraction{.001}

% ── Standalone title page ─────────────────────────────────────────────────────
\newpage
\thispagestyle{empty}

\begin{center}
{\LARGE \textbf{DIGOM: Disentangled Interaction Graphs for Outcome prediction in Multimodal healthcare}}

\vspace{1.5cm}

{\large
Hakima Laribi$^{1,*}$,
Olivier Lefebvre$^{1}$,
Dan Poenaru$^{2,3}$,
Martin Vallières$^{4,1}$
}

\vspace{1cm}

{\normalsize
$^{1}$Department of Computer Science, Université de Sherbrooke,\\
2500 Boulevard de l'Université, Sherbrooke, Québec, J1K 2R1, Canada\\[0.4cm]

$^{2}$Department of Pediatric Surgery, Montreal Children's Hospital,\\
1001 Boulevard Décarie, Montreal, Québec, H4A 3J1, Canada\\[0.4cm]

$^{3}$Centre for Outcomes Research and Evaluation (CORE),\\
Research Institute of the McGill University Health Centre,\\
5252 Boulevard de Maisonneuve Ouest, Montreal, Québec, H4A 3S5, Canada\\[0.4cm]

$^{4}$Medical Physics Unit, Department of Oncology, McGill University,\\
1001 Boulevard Décarie, Montreal, Québec, H4A 3J1, Canada
}

\vspace{1cm}

{\normalsize
$^{*}$\textbf{Corresponding author:}\\
Hakima Laribi\\
Department of Computer Science, Université de Sherbrooke\\
2500 Boulevard de l'Université, Sherbrooke, Québec, J1K 2R1, Canada\\
Email: \texttt{hakima.laribi@usherbrooke.ca}
}

\end{center}

\newpage
\shorttitle{DIGOM: Disentangled Interaction Graphs for Outcome prediction in Multimodal healthcare}    

\shortauthors{}  

\title[mode = title]{DIGOM: Disentangled Interaction Graphs for Outcome prediction in Multimodal healthcare}

\author[1]{Hakima Laribi}[orcid=0009-0005-7773-9801]
\cormark[1]
\ead{hakima.laribi@usherbrooke.ca}
\credit{Conceptualization, Data curation, Formal analysis, Investigation, Methodology, Software, Validation, Visualization, Project administration, Writing -- Original Draft}

\affiliation[1]{organization={Department of Computer Science, Université de Sherbrooke},
            city={Sherbrooke},
            country={Canada}}

\author[1]{Olivier Lefebvre}[orcid=0009-0000-8043-2185]
\credit{Investigation, Validation, Writing -- Review \& Editing}

\author[2,3]{Dan Poenaru}[orcid=0000-0002-6267-6140]
\credit{Conceptualization, Methodology, Validation, Supervision, Writing -- Review \& Editing}

\affiliation[2]{organization={Department of Pediatric Surgery, Montreal Children's Hospital},
            city={Montreal},
            country={Canada}}

\affiliation[3]{organization={Centre for Outcomes Research and Evaluation (CORE), Research Institute of the McGill University Health Centre},
            city={Montreal},
            country={Canada}}

\author[4,1]{Martin Vallières}[orcid=0000-0001-7639-8172]
\credit{Conceptualization, Funding acquisition, Methodology, Project administration, Resources, Supervision, Validation, Writing -- Review \& Editing}

\affiliation[4]{organization={Medical Physics Unit, Department of Oncology, McGill University},
            city={Montreal},
            country={Canada}}

\cortext[1]{Corresponding author}


\begin{abstract}
Multimodal healthcare data contain relational information that extends beyond individual patient observations. Effective clinical outcome prediction therefore requires not only integrating information from multiple modalities, but also modeling the relationships between them. Standard fusion approaches do not explicitly capture these interactions, while existing graph-based approaches typically model only a subset of them. We propose DIGOM (Disentangled Interaction Graphs for Outcome prediction in Multimodal healthcare), a graph-based framework that jointly models three complementary interaction types: \textit{Intra-patient cross-modal}, \textit{Cross-patient same-modal}, and \textit{Cross-patient cross-modal} interactions. DIGOM learns a sparse multiplex graph with disentangled views, each encoding interactions between pairs of modalities across patients. Discrete adjacency matrices with patient-adaptive connectivity are sampled via a differentiable mechanism, enabling end-to-end learning and allowing the model to select informative relationships for each patient without imposing a fixed number of neighbors. A regularization objective maintains overall graph sparsity, while message passing across the disentangled views integrates this relational context into patient representations for outcome prediction. We evaluate DIGOM on two multimodal healthcare datasets, MIMIC-IV and NIH-TB, across five clinical prediction tasks. DIGOM achieves the highest AUROC on all five tasks while using substantially fewer connections than existing multi-interaction graph approaches. Ablation studies reveal that the contribution of each interaction type varies across tasks, highlighting the need to adapt relational structures to each task. DIGOM also demonstrates greater robustness to missing modalities, suggesting that the learned relationships provide complementary information when some modalities are unavailable. Post-hoc analyses further reveal task- and outcome-specific connectivity patterns in the learned graphs.
\end{abstract}


\begin{highlights}
    \item DIGOM jointly models three multimodal interaction types via a disentangled graph.
    \item Patient-adaptive sparse connectivity is learned via differentiable discrete sampling.
    \item Message passing across disentangled graph views enriches patient representations.
    \item DIGOM achieves SOTA on five clinical tasks across two healthcare datasets.
\end{highlights}

%\nocite{*}

% Keywords
\begin{keywords}
Multimodal fusion \sep Clinical outcome prediction\sep Graph learning\sep Multiplex graphs\sep Disentangled graphs
\end{keywords}
\maketitle

% Main text
\section{Introduction}\label{sec:introduction}
The growing availability of digital healthcare data creates new opportunities for artificial intelligence to improve patient care~\cite{tang2024harnessing}. Depending on a patient's condition and the available hospital resources, different types of information may be collected during their hospital stay, including demographic information, clinical notes, vital signs, laboratory test results, medical imaging, and genomic profiles. These sources of information, referred to as modalities, each provide a partial description of a patient's condition, while together, they offer a more complete view of their health status~\cite{soenksen2022integrated}. This naturally reflects the way clinicians reason in practice, where they integrate complementary sources of information before making diagnostic or prognostic decisions. Consequently, multimodal learning has become an active area of research in medical artificial intelligence, frequently demonstrating improved predictive performance compared with unimodal approaches across a wide range of clinical applications~\cite{yang2026multimodal, vavekanand2026multimodal}.

Despite this promise, effectively integrating multimodal healthcare data remains a major challenge. Standard multimodal fusion approaches are typically classified into three categories: early fusion, which concatenates modalities before learning; joint fusion, which combines latent representations learned from each modality; and late fusion, which aggregates predictions generated independently by each modality~\cite{krones2025review}. Although effective in many settings, these approaches ultimately aggregate modality-specific information into a single representation or prediction without modeling relationships between individual observations across modalities. Consequently, they may fail to capture how observations from different modalities complement or reinforce each other, limiting their ability to leverage fine-grained relationships in multimodal healthcare data.

However, medical decision-making rarely consists of simply aggregating independent observations. Instead, clinicians interpret each finding in relation to other clinical findings, as the significance of one observation often depends on the context provided by others~\cite{croskerry2009universal}. For example, bilateral pulmonary infiltrates on chest imaging become suggestive of severe pneumonia when accompanied by elevated temperature and inflammatory markers. Likewise, physicians routinely compare current patients with previously encountered similar cases. A patient presenting with a chest radiograph or laboratory profile similar to previously observed cases with poor outcomes may provide valuable prognostic information~\cite{lee2015personalized}. Once similar patients are identified, their remaining clinical profiles, including laboratory measurements, vital signs, procedures, and treatments, can provide additional information for understanding the current patient and anticipating their clinical course~\cite{gottlieb2013method}. As such, beyond comparing the same modality across patients, clinicians also leverage knowledge acquired from previous cases to relate findings across modalities, anticipating which abnormalities are likely to emerge next or which interventions may become necessary. Importantly, some decisions may rely primarily on complementary findings within a patient, while others may benefit more from information provided by similar patients or from other modalities.

\begin{figure*} 
\centering 
\begin{subfigure}[b]{0.5\textwidth} 
         \centering 
           \includegraphics[width=\linewidth]{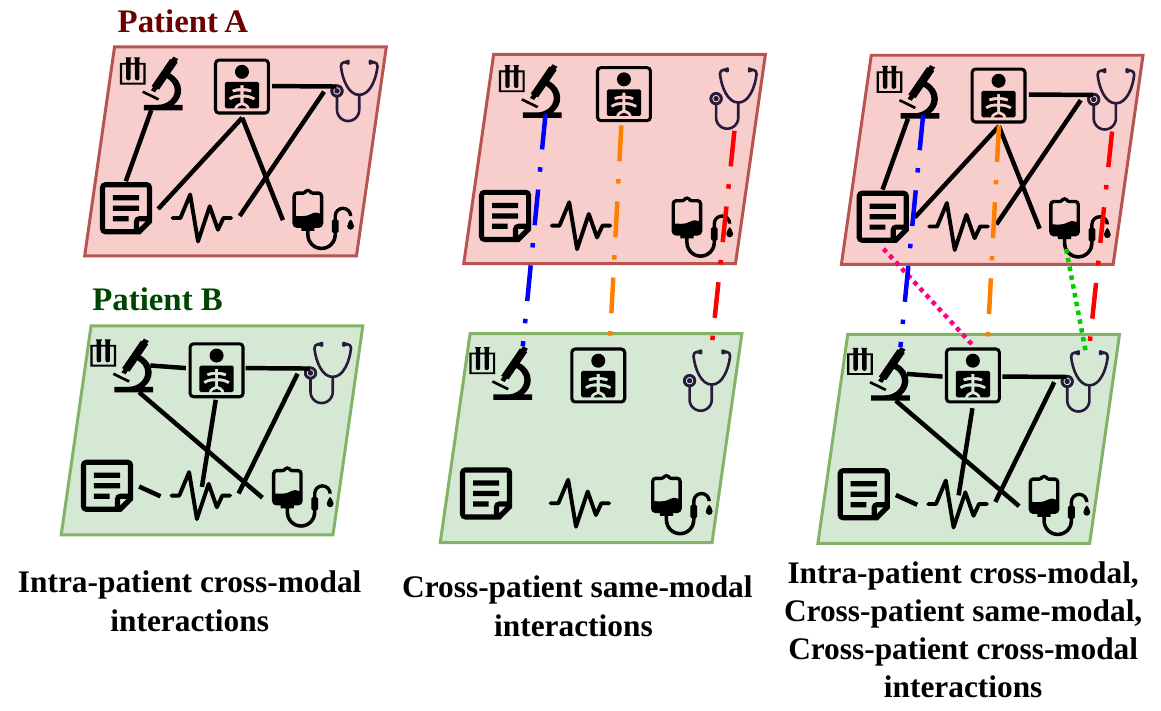}
         \label{subfig:interact_type} 
         \caption{}
     \end{subfigure} 
     \hspace{1.3em}
\begin{subfigure}[b]{0.43\textwidth} 
         \centering 
           \includegraphics[width=\linewidth]{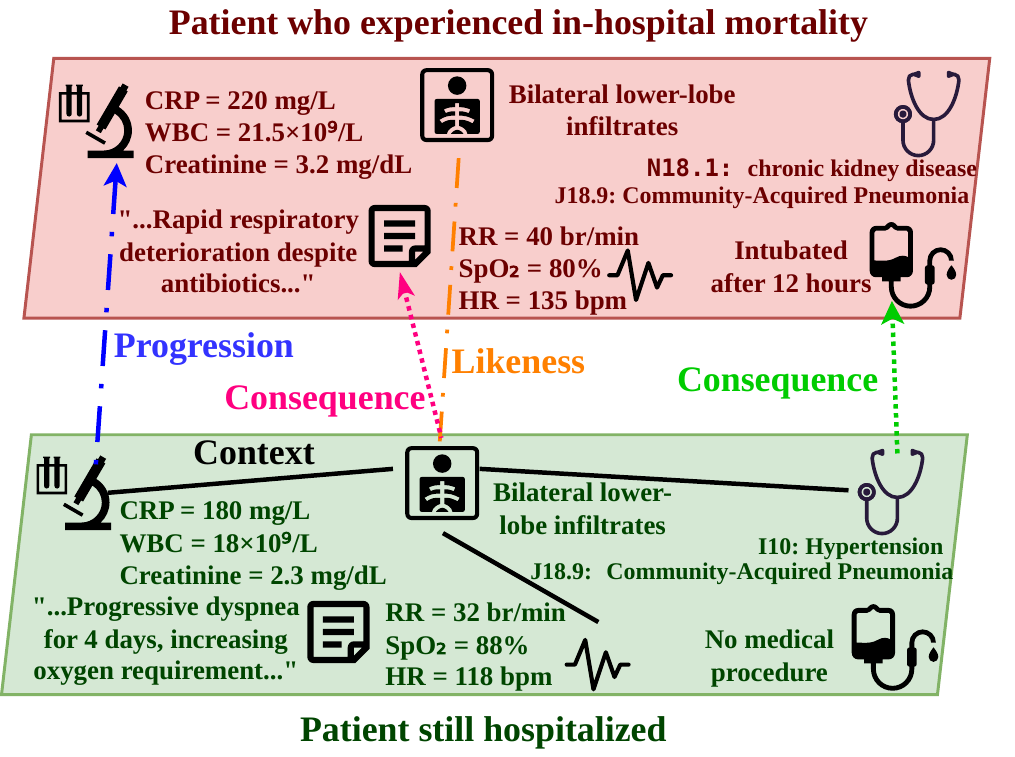}
         \label{subfig:ex_interact} 
         \caption{}
     \end{subfigure} 
\caption{Illustration of the three interaction types modeled by DIGOM and their role in clinical reasoning. (a) Existing methods typically capture only one of Intra-patient cross-modal or Cross-patient same-modal interactions, whereas DIGOM jointly models both along with Cross-patient cross-modal interactions, capturing a richer relational structure across patients and modalities. (b) A clinical case illustrating how a newly admitted patient (green) can benefit from information from a previously seen patient (red) with similar clinical findings who experienced in-hospital mortality. Shared observations help identify the related patient, while their remaining clinical information and outcome provide additional context for understanding the current patient's condition and anticipating their outcome.}
\label{fig:interactions}
\end{figure*}

Based on these observations, we categorize clinically relevant interactions into three complementary forms: (i) \textit{Intra-patient cross-modal} interactions, which capture relationships between different modalities within the same patient; (ii) \textit{Cross-patient same-modal} interactions, which capture relationships involving the same modality across different patients; and (iii) \textit{Cross-patient cross-modal} interactions, which capture relationships between different modalities across different patients. Figure~\ref{fig:interactions} illustrates these three categories of interactions, which constitute the focus of this work. Note that we use \textit{same-modal} to refer to relationships involving the same modality across patients, rather than \textit{intra-modal}, which conventionally denotes relationships between variables within a single modality.

Graph-based methods provide a natural mechanism for modeling such relational reasoning by representing patients or their modalities as nodes and their relationships as edges, through which Graph Neural Networks (GNNs)~\cite{scarselli2008graph} propagate information for prediction. For fusion, existing approaches generally follow one of two paradigms. The first constructs patient graphs, where different modalities belonging to the same patient are connected to capture Intra-patient cross-modal interactions~\cite{hou2023hybrid, al2025multimodal}. The second builds population graphs, connecting patient modalities according to feature similarity or shared clinical characteristics to leverage Cross-patient same-modal interactions~\cite{zhang2022m3care,tang2023predicting, kim2023heterogeneous}. Although both paradigms have demonstrated promising results, they have largely evolved independently. Patient graphs ignore potentially informative cohort-level relationships, whereas population graphs overlook the rich interactions that exist between modalities within an individual patient. Consequently, neither paradigm captures the full range of relationships described above.

More recently,~\cite{d2023maxcorrmgnn} proposed jointly modeling the three intra-patient and cross-patient interactions defined in Figure~\ref{fig:interactions}, bringing graph-based fusion closer to the clinical reasoning described above and demonstrating promising results. However, although a sparsity mechanism was introduced to control the graph connectivity, the resulting graphs remain relatively dense because the method does not explicitly control the number of retained edges. Such a large number of relationships increases computational cost, limits scalability to larger cohorts~\cite{chen2020scalable}, may promote over-smoothing~\cite{chen2020measuring}, and reduces the interpretability of the learned graph. Furthermore, this sparsity mechanism is uniform across all patients, where a single threshold determines which relationships are retained, regardless of the nodes involved. A natural extension is to make graph connectivity patient-adaptive, because clinically less complex cases may require only a few informative relationships, whereas more complex cases may benefit from integrating evidence from a larger number of related patients.

Beyond graph construction, little attention has been devoted to understanding the contribution of each interaction type to prediction accuracy. Furthermore, multimodal healthcare data are frequently incomplete, as imaging studies, laboratory measurements, or other clinical observations may be unavailable for some patients~\cite{le2025multimodal}. However, the ability of the learned connections to compensate for missing information by propagating evidence from related modalities remains underexplored.

In this work, we hypothesize that identifying a sparse, patient-specific subset of informative intra- and cross-patient relationships across modalities can improve predictive performance while reducing computational complexity and improving scalability. We further hypothesize that jointly modeling these complementary interactions can improve robustness to missing modalities by allowing information to propagate across related modalities. To test these hypotheses, we propose \textbf{DIGOM} (\textbf{\underline{D}}isentangled \textbf{\underline{I}}nteraction \textbf{\underline{G}}raphs for \textbf{\underline{O}}utcome prediction in \textbf{\underline{M}}ultimodal healthcare), a graph-based framework that jointly models Intra-patient cross-modal, Cross-patient same-modal, and Cross-patient cross-modal interactions through a learned sparse multiplex graph. Each view of the graph encodes directed patient-to-patient relationships between a pair of modalities. This disentangled design allows each interaction type to be isolated to assess its contribution to predictive performance. Rather than assuming that all possible connections contribute equally, DIGOM retains only a subset of candidate relationships, whose number can vary across patients, resulting in a sparser and more analyzable graph.

Our contributions are summarized as follows:
\begin{itemize}
\item We propose DIGOM, a disentangled graph-based framework that explicitly models complementary intra- and cross-patient interactions across modalities in multimodal healthcare data.
\item We introduce a patient-adaptive sparse graph learning strategy that identifies informative relationships for each patient, rather than imposing a fixed connectivity pattern across the cohort.
\item We provide a comprehensive benchmark of existing multimodal fusion and graph-based approaches across two multimodal healthcare datasets and five clinical prediction tasks, in which DIGOM achieves the highest mean AUROC on all five tasks.
\item We evaluate the robustness of the learned representations to missing modalities, showing that multimodal interactions can compensate for unavailable modalities.
\item We perform ablation and post-hoc analyses of the disentangled graph structure to quantify the contribution of each interaction type and analyze learned connectivity patterns across clinical prediction tasks and patient subgroups.
\end{itemize}

The remainder of this paper is organized as follows. Section~\ref{sec:relatedwork} reviews related work on multimodal learning in healthcare and graph structure learning. Section~\ref{sec:methods} introduces the DIGOM framework, detailing the patient-adaptive sparse graph learning module, the multimodal multi-interaction message passing module, and the training procedure. Section~\ref{sec:setup} describes the datasets, baselines, and evaluation protocol. Section~\ref{sec:results} presents the results, including performance comparisons, interaction type ablations, robustness to missing modalities, graph sparsity analysis, and post-hoc connectivity analyses. Section~\ref{sec:discussion} discusses the findings, limitations, and directions for future work.

\section{Related work}\label{sec:relatedwork}
\subsection{Multimodal learning in healthcare}
Standard multimodal fusion approaches combine heterogeneous healthcare data through early, joint, or late fusion strategies. These methods have demonstrated improved performance over unimodal approaches across various medical applications~\cite{soenksen2022integrated, subramanian2020multimodal}, but they primarily aggregate multimodal data and do not explicitly model the underlying relationships between modalities. More recent attention-based fusion approaches have enabled the modeling of multimodal interactions through cross-attention mechanisms and have achieved strong performance across multiple clinical tasks~\cite{susetzky2025holistic, lu2024hierarchical}. However, they typically produce dense interaction patterns, where every modality contributes to the final representation, rather than identifying a sparse subset of informative relationships. Moreover, these interactions are typically limited to modalities within the same patient and do not consider relationships across patients. Alternatively, contrastive learning approaches have been proposed to align heterogeneous modalities within shared representation spaces~\cite{wang2022medclip, zhou2022generalized}. While effective for learning modality-invariant representations, these methods primarily focus on improving patient-level representations rather than capturing relational interactions between modalities or patients. More recently, multimodal large language models (MLLMs) have emerged as a promising paradigm for integrating heterogeneous medical data, demonstrating strong capabilities in clinical reasoning and generative tasks~\cite{alsaad2024multimodal}. However, their application to supervised clinical tasks remains an active area of investigation, with no clear consensus regarding their superiority over conventional specialized machine learning models~\cite{brown2025large, vishwanath2026general}.

Due to their ability to model relationships between entities, graph-based methods have been increasingly used in healthcare applications along with GNNs. Early approaches constructed population graphs where each node represents a patient with multimodal features aggregated into a single representation, while edges capture similarities based on one or more patient characteristics~\cite{parisot2017spectral, cosmo2020latent}. These methods represent each patient as a single multimodal node, limiting interactions between individual modalities. Subsequent works shifted toward modality-level graph representations to allow for multimodal interactions. Several studies used graphs to model intra-patient interactions by connecting different modalities belonging to the same patient~\cite{al2025multimodal,wang2025multimodal}.~\cite{hou2023hybrid} modeled intra-modal relationships using graphs and Intra-patient cross-modal relationships via attention mechanism, while~\cite{zheng2022multi} first models Intra-patient cross-modal interactions with Transformers before constructing population graphs to capture cross-patient relationships. Conversely, another line of research focused on cross-patient relationships through multiplex graphs, where each graph view represents patient similarities according to a specific modality~\cite{zhang2022m3care,tang2023predicting} or a group of features~\cite{kim2023heterogeneous}. Patient representations are then obtained by aggregating information across these multiple graph views. More recently,~\cite{d2022fusing, d2024fusing} proposed to jointly model intra- and cross-patient interactions. Their initial approach connected each node to its occurrences across multiplex graph views, allowing Intra-patient cross-modal interactions as well as Cross-patient cross-modal interactions through multi-hop message passing. They later generalized this formulation by allowing direct connections between nodes from any graph view~\cite{d2023maxcorrmgnn}, enabling explicit modeling of Intra-patient cross-modal, Cross-patient same-modal, and Cross-patient cross-modal interactions within a single graph. This unified formulation demonstrated improved performance over previous graph-based approaches for tuberculosis outcome prediction. However, the resulting interaction graphs remain densely connected, limiting scalability, computational efficiency, and interpretability. Moreover, the contribution of the different interaction types and their ability to compensate for missing modalities remain unexplored.

\subsection{Graph structure learning}
In many applications, the graph structure describing relationships between entities is not readily available and must be inferred from data. Static approaches construct a fixed graph prior to training using either domain knowledge~\cite{xu2022counterfactual, cai2022hypergraph} or pairwise similarities~\cite{parisot2017spectral, tang2023predicting, kim2023heterogeneous} computed from raw features, followed by neighbor selection strategies such as \textit{k}-nearest neighbors (KNN) or thresholding. These predefined graphs remain fixed throughout training and may therefore fail to capture task-specific relationships. Alternatively, dynamic graph construction methods update the graph throughout training using similarity-based neighbor selection on latent node representations~\cite{wang2022contrastive, wang2019dynamic}. Although the graph evolves with the learned representations, its construction is not directly optimized through the prediction objective. More recent latent graph learning approaches jointly optimize graph structures and model parameters in an end-to-end manner, allowing the task loss to directly shape graph connectivity. However, maintaining sparse graphs while preserving differentiability remains a major challenge, as gradient-based optimization typically relies on continuous adjacency matrices, which correspond to fully connected graphs. Several strategies have been proposed to alleviate this issue.~\cite{franceschi2019learning} jointly optimized discrete adjacency matrices and model parameters through bi-level optimization. This typically prevents fully connected graphs but does not explicitly control graph sparsity. Other methods enforce sparse connectivity using differentiable implementations of KNN or thresholding~\cite{fatemi2021slaps, kazi2022differentiable, zheng2022multi, d2023maxcorrmgnn}; however, they rely on a fixed number of neighbors or a global threshold shared across all nodes. More recently,~\cite{zaripova2023graph} encouraged sparse graphs by regularizing the node degree distribution around a learned target parameter, allowing nodes to have varying degrees while preserving a desired global connectivity pattern. Nevertheless, the method still relies on a continuous adjacency matrix, where many weak connections may persist despite a low target node degree. Consequently, learning sparse, task-specific graph structures with node-adaptive connectivity remains an open challenge.

\section{Methods}\label{sec:methods}
\subsection{Problem formulation}
\subsubsection{Notation}
We consider a multimodal healthcare dataset consisting of $N$ patients and $M$ modalities. Each patient $P_i$ is represented by their features $X_i = [x_i^1, x_i^2, \dots, x_i^M]$, with $x_i^k \in \mathbb{R}^{d_{m_k}}$ denoting the vector representation of modality $m_k$ for $P_i$. Our goal is to learn a multimodal multi-interaction graph that captures interactions between modalities both within and across patients. We formulate this graph as a disentangled multiplex bipartite graph, where each view corresponds to a directed interaction from a source modality $m_k$ to a target modality $m_l$. This formulation allows the same pair of patients to be connected through different modality-specific relationships, while preserving the distinction between interaction types.

Let $\mathcal{G}=(\mathcal{V},\mathcal{E})$ denote the resulting disentangled multiplex bipartite graph. For each modality $m_k$, we define a source-side and a target-side copy of the patient set as $\mathcal{V}_{\mathrm{src}}^k = \{(P_i, m_k)_{\mathrm{src}} : i \in [N]\}$ and $\mathcal{V}_{\mathrm{tgt}}^k = \{(P_i, m_k)_{\mathrm{tgt}} : i \in [N]\}$, where $(P_i, m_k)_{\mathrm{src}}$ and $(P_i, m_k)_{\mathrm{tgt}}$ are the source- and target-role copies of the node representing modality $m_k$ of patient $P_i$. Letting $\mathcal{V}_{\mathrm{src}} = \bigcup_{k=1}^M \mathcal{V}_{\mathrm{src}}^k$ and $\mathcal{V}_{\mathrm{tgt}} = \bigcup_{k=1}^M \mathcal{V}_{\mathrm{tgt}}^k$, the vertex set is partitioned into two disjoint sets $\mathcal{V} = \mathcal{V}_{\mathrm{src}} \cup \mathcal{V}_{\mathrm{tgt}}$. The edge set consists of $M^2$ relation types: $\mathcal{E} = \bigcup_{k=1}^{M}\bigcup_{l=1}^{M}\mathcal{E}^{k,l}$, where $\mathcal{E}^{k,l} \subseteq \mathcal{V}^{k}_{\mathrm{src}}\times\mathcal{V}^{l}_{\mathrm{tgt}}$ defines a view for propagating information from modality $m_k$ of source patients to modality $m_l$ of target patients. Although the node set differs across views, the graph remains a multiplex graph, as the nodes of every view correspond to the same set of patients and differ only in the modality they represent. Each view is associated with an adjacency matrix $A^{k,l} \in \{0,1\}^{N \times N}$, where $A^{k,l}_{ij}=1$ indicates an active edge from modality $m_k$ of patient $P_i$ to modality $m_l$ of patient $P_j$, and $0$ otherwise. Following the principle of disentangled graph representations~\cite{ma2019disentangled}, the edge set is partitioned into $M^2$ disjoint views $\{\mathcal{E}^{k,l}\}_{k,l=1}^{M}$, each encoding a distinct directed relation between a pair of modalities. We define the neighborhood of a target node $(P_i, m_l)_{\mathrm{tgt}}$ in view $\mathcal{E}^{k,l}$ as $\mathcal{N}^{k,l}(i) = \{(P_j, m_k)_{\mathrm{src}} : A^{k,l}_{ji}=1,\ j \in [N]\}$.

\subsubsection{Overview} 
\begin{figure*} 
\centering 
\includegraphics[width=\linewidth]{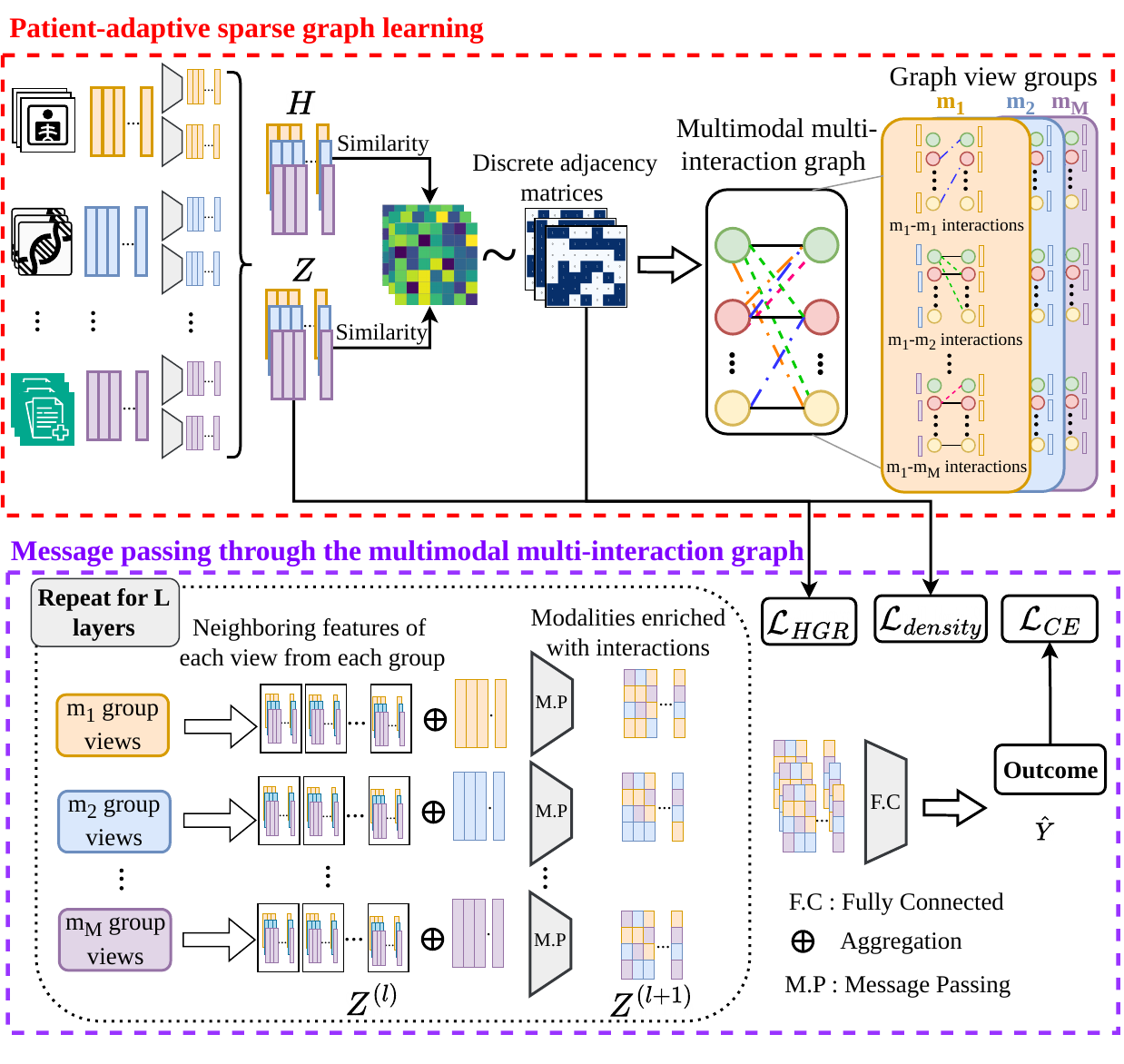}
\caption{Overview of the DIGOM framework. (\textit{Top}) The patient-adaptive sparse graph learning module produces two latent projections $H$ and $Z$ for each modality, computes pairwise similarity matrices, and samples discrete adjacency matrices to construct a sparse multimodal multi-interaction graph organized into $M^2$ disentangled views, each encoding directed interactions from a source modality to a target modality across all patients. Views are grouped by target modality, such that each group contains all incoming interactions for a given modality. (\textit{Bottom}) The multimodal multi-interaction message passing module propagates information across the learned graph for $L$ layers. At each layer, representations of target modality $m_l$ are updated by aggregating neighboring representations from all $M$ source modality views and combining them with the node's own representation. The resulting enriched modality representations are concatenated and passed through a fully connected layer to produce the final outcome prediction $\hat{Y}$. The model is jointly optimized using the cross-entropy loss $\mathcal{L}_{\mathrm{CE}}$, the soft-HGR correlation loss $\mathcal{L}_{\mathrm{sHGR}}$, and the density regularization loss $\mathcal{L}_{\mathrm{density}}$.}
\label{fig:architecture}
\end{figure*}

Our framework, DIGOM, consists of two main components (Figure~\ref{fig:architecture}): (i) a \textit{patient-adaptive sparse graph learning} module, which learns latent projections for each modality and constructs a sparse multimodal multi-interaction graph by sampling edges from learned pairwise similarities; and (ii) a \textit{multimodal multi-interaction message passing} module, which propagates information across the learned graph to obtain enriched, context-aware modality representations and combines them for patient-level outcome prediction. We describe each component in turn below.

\subsection{Patient-adaptive sparse graph learning}
In this section, we describe how the multimodal multi-interaction graph $\mathcal{G}$ introduced above is learned, as illustrated in Figure~\ref{fig:architecture}. The graph learning process consists of four steps: (i) projecting each modality into a common latent space, (ii) computing pairwise similarities between modality representations, (iii) sampling discrete adjacency matrices from the resulting similarities, and (iv) controlling graph density.

For each modality, we first learn two distinct projections that map all modalities into a common low-dimensional latent space, in which pairwise similarities can be meaningfully compared. Specifically, for modality $m_k$ of patient $P_i$, we compute two latent representations of the modality features $x_i^k$:
\begin{equation}
h_i^k = f_\phi^k(x_i^k), \qquad z_i^k = f_\theta^k(x_i^k),
\end{equation}
where $f_\phi^k$ and $f_\theta^k$ are distinct modality-specific projection networks mapping into a shared $d$-dimensional space. We denote by $H^k$ and $Z^k$ the matrices stacking all $h_i^k$ and $z_i^k$, respectively, which are used jointly to compute pairwise similarities between modality representations and thereby determine graph connectivity. We use two projections rather than a single shared one to avoid a trivial solution, in which representations of the same modality from different patients appear artificially similar simply because they are produced by the same network and share the same output distribution. Moreover, while $H$ is used only for similarity computation, $Z$ also serves as the node representation during message passing. The two projections are therefore optimized for different roles, which prevents them from collapsing into the same representation.

For each ordered modality pair $(k,l)$, we compute a similarity matrix $S^{k,l} \in \mathbb{R}^{N \times N}$ from the source projections $H^k$ and the target projections $Z^l$:
\begin{equation}
s^{k,l}_{ij} = \frac{{z_j^l}^\top h_i^k}{\lVert z_j^l \rVert \, \lVert h_i^k \rVert}
\end{equation}

This similarity matrix could, in principle, be used directly as an adjacency matrix. However, such an adjacency matrix would result in a fully connected graph across all $M^2$ relation types, which scales poorly with the number of modalities and patients and precludes any meaningful interpretation of the learned structure. We therefore convert each similarity matrix into a \emph{discrete} adjacency matrix $A^{k,l} \in {\left\{0, 1\right\}}^{N \times N}$. The similarity score is first scaled by a temperature hyperparameter $\tau$ and passed through a sigmoid to obtain the parameter of a Bernoulli distribution over the existence of edge $e_{ij}^{k,l}$:
\begin{equation}
p^{k,l}_{ij} = \sigma\!\left(\frac{s^{k,l}_{ij}}{\tau}\right)
\end{equation}
where $\sigma$ denotes the sigmoid function. The temperature $\tau$ controls the sharpness of the resulting probability distribution, where lower temperatures push probabilities closer to 0 or 1. This reduces the entropy of the sampling distribution and encourages the model to make decisive connectivity choices rather than sampling a large number of weak relationships. For each modality pair, an adjacency matrix is then sampled according to these probabilities. Specifically, the presence of an edge between modality $m_k$ of patient $P_i$ and modality $m_l$ of patient $P_j$ is modeled as a binary random variable $a^{k,l}_{ij}$, drawn from a Bernoulli distribution with parameter $p^{k,l}_{ij}$. In other words, the edge is retained ($a^{k,l}_{ij} = 1$) with probability $p^{k,l}_{ij}$ and discarded ($a^{k,l}_{ij} = 0$) otherwise:
\begin{equation}
\label{eq:bernoulli}
a^{k,l}_{ij} \sim \operatorname{Bernoulli}\!\left(p^{k,l}_{ij}\right)
\end{equation}

However, sampling from a Bernoulli distribution is not differentiable, and directly incorporating it into the graph learning process would break the gradient flow. To preserve end-to-end learning while still operating on a discrete adjacency matrix, we use a Straight-Through (ST) estimator~\cite{bengio2013estimating} on top of the Gumbel-Softmax reparameterization trick~\cite{jang2016categorical}. Concretely, we use Gumbel-Softmax to compute a continuous relaxation of each sampled adjacency matrix entry as follows:
\begin{equation}
{a^{\prime}}_{ij}^{k,l} = \frac{\exp\!\left(\dfrac{\log(p_{ij}^{k,l}) + g_1}{\tau_{\mathrm{gs}}}\right)}{\exp\!\left(\dfrac{\log(p_{ij}^{k,l}) + g_1}{\tau_{\mathrm{gs}}}\right) + \exp\!\left(\dfrac{\log(1-p_{ij}^{k,l}) + g_0}{\tau_{\mathrm{gs}}}\right)}
\label{eq:gumbel}
\end{equation}
with
\begin{equation}
g_c = -\log(-\log(u_c)) \sim \mathrm{Gumbel}(0,1),
\end{equation}
where $u_c \sim \mathrm{Uniform}(0,1)$ and $c \in \{0, 1\}$, so that $g_0$ and $g_1$ are independent Gumbel noise samples, and $\tau_{\mathrm{gs}}$ is the Gumbel-Softmax temperature. In the forward pass, we use the discrete adjacency matrix computed in Equation~\ref{eq:bernoulli} so that DIGOM operates on a discrete graph. In the backward pass, the ST estimator propagates gradients directly through the continuous relaxation ${a^{\prime}}_{ij}^{k,l}$. This preserves end-to-end learning while preventing the adjacency matrix from collapsing into a dense, continuous approximation.
Unlike KNN- or threshold-based approaches, this formulation does not impose a shared number of neighbors or a shared cutoff on every node. However, without any additional constraint, independently sampling each edge from $p^{k,l}_{ij}$ may still produce unnecessarily dense graphs. We therefore introduce a density loss that regularizes the overall connectivity of the sampled graphs, without imposing a strict, uniform neighbor count across nodes:
\begin{equation}
\label{eq:density}
\mathcal{L}_{\text{density}} = \sum_{(k,l)} \left(\frac{1}{N^2}\sum_{(i,j)} a^{k,l}_{ij}  -  \rho \right)^{2}
\end{equation}
where $\rho\in(0,1)$ specifies the desired average density of each graph view. Importantly, this regularization constrains the average connectivity rather than the number of neighbors of each individual node. The number of connections can therefore vary across patients, allowing the model to allocate more connections to patients for whom additional contextual information is useful and fewer to patients who can be adequately represented using fewer relationships. This provides sparsity at the graph level while allowing patient-adaptive connectivity.

Once every relation-specific adjacency matrix $A^{k,l}$ has been sampled, the resulting structure instantiates the disentangled multiplex bipartite graph $\mathcal{G} = (\mathcal{V}, \mathcal{E})$ defined above. The $M^2$ views jointly encode the three interaction categories considered in this work: Intra-patient cross-modal interactions ($k\neq l, i=j$), Cross-patient same-modal interactions ($k=l, i\neq j$), and Cross-patient cross-modal interactions ($k\neq l, i\neq j$).

\subsection{Multimodal multi-interaction message passing for outcome prediction}
The learned multiplex graph $\mathcal{G}$ provides the structure through which modality representations exchange information across patients and modalities. We use the $Z$ projections as the initial node representations for message passing. If representations from different modalities  $\{Z^k : k \in [M]\}$ are poorly aligned, information propagated between them may be less exploitable. Conversely, if the representations are highly redundant, message passing may carry less complementary information. To encourage cross-modal alignment while regularizing the covariance structure of the learned representations, we supervise the $Z$ projections using a soft Hirschfeld--Gebelein--R\'enyi (HGR) correlation objective~\cite{wang2019efficient}. We first define the soft-HGR correlation between the projections of two distinct modalities $m_k$ and $m_l$ as:
\begin{equation}
r_{\text{sHGR}}(Z^k, Z^l) = \frac{1}{N-1}\sum_{i=1}^{N} {z_i^k}^\top z_i^l \; - \; \frac{1}{2}\operatorname{Tr}\!\left(\operatorname{Cov}(Z^k)\operatorname{Cov}(Z^l)\right)
\end{equation}
and define the soft-HGR loss as its negative, averaged over all ordered pairs of distinct modalities:
\begin{equation}
\mathcal{L}_{\text{sHGR}} = -\frac{1}{M(M-1)} \sum_{\substack{k,l \\ k \neq l}} r_{\text{sHGR}}(Z^k, Z^l) 
\end{equation}

For each target modality $m_l$, the disentangled multiplex graph provides a group of $M$ views in which $m_l$ acts as the target modality and receives information from every source modality $m_k$ through the learned graph connectivity: $\mathcal{E}^{l} = \bigcup_{k=1}^{M}\mathcal{E}^{k,l}$. We use the Graph Isomorphism Network (GIN)~\cite{xu2018powerful} as the message-passing function and adapt it to this multiplex bipartite setting. First, for each source modality $m_k$, neighbor representations are aggregated within the corresponding view $\mathcal{E}^{k,l}$ for a patient $P_i$:
\begin{equation}
{e_i^{k \rightarrow l}}^{(t)} =
\frac{1}{1 + |\mathcal{N}^{k,l}(i)|}
\sum_{j \in \mathcal{N}^{k,l}(i)}
{z_j^k}^{(t)}
\end{equation}
where $\mathcal{N}^{k,l}(i)$ denotes the set of source patients connected to patient $P_i$ through the relation from source modality $m_k$ to target modality $m_l$, and ${z_j^k}^{(0)} = z_j^k$. While the original GIN uses sum aggregation to preserve its injective property, directly summing messages in our setting makes their magnitude highly dependent on the number of neighbors, which can affect learning. This is particularly important for the learned graph $\mathcal{G}$, where node degrees can vary substantially across patients. We therefore normalize the aggregated message by the neighborhood size of each target node $(P_i,m_l)_{\mathrm{tgt}}$ before combining information across views. The $M$ aggregated messages are then summed across source modality views, combined with the target node's own representation, and passed through the GIN update function:
\begin{equation}
{z_i^l}^{(t+1)}= f_\psi\!\left( (1+\epsilon){z_i^l}^{(t)} + \sum_{k=1}^{M} {e_i^{k \rightarrow l}}^{(t)} \right)
\end{equation}
where $\epsilon$ is a scalar weighting the node's own representation. This update is applied to every modality $m_l$ and patient $P_i$ and repeated for $L$ message-passing layers.

Following $L$ layers of message passing, the enriched modality representations $\{{z_i^l}^{(L)}\}_{l=1}^{M}$ of patient $P_i$ are concatenated to form a joint patient representation. This representation is then passed through a prediction head $g$ to obtain the final clinical outcome prediction:
\begin{equation}
\hat{y}_i =
\sigma_t\!\left(
g\!\left(
\operatorname{Concat}
\left(
{z_i^1}^{(L)},\ldots,{z_i^M}^{(L)}
\right)
\right)
\right)
\end{equation}
where $\sigma_t$ denotes either a sigmoid or a softmax function, depending on the prediction task.

\subsection{Model optimization}
Training begins with a warm-up phase of $L_{\text{warmup}}$ epochs, during which only the soft-HGR loss and the density loss are optimized, allowing the latent graph structure to stabilize before it is used for downstream prediction:
\begin{equation}
\mathcal{L}_{\text{warmup}} = (1- \lambda_1) \mathcal{L}_{\mathrm{sHGR}} + \lambda_1 \, \mathcal{L}_{\mathrm{density}}
\end{equation}
where $\lambda_1 \in (0,1)$.
 
Following the warm-up phase, the full training objective is optimized. We use the cross-entropy loss to supervise the outcome predictions, instantiated as binary or categorical cross-entropy depending on the clinical task. The full objective is:
\begin{equation}
\mathcal{L} = (1- \lambda_2)  \mathcal{L}_{\mathrm{CE}} \; + \; \lambda_2 \left((1- \lambda_1) \mathcal{L}_{\mathrm{sHGR}} + \lambda_1 \, \mathcal{L}_{\mathrm{density}}\right)
\end{equation}
where $\lambda_2 \in (0,1)$.

\subsubsection{Inductive learning}
To ensure that our framework generalizes to unseen patients at inference time, we adopt an inductive learning setting throughout training and evaluation. During training, the graph is constructed exclusively from patients in the training set. During validation, the learned model parameters are frozen and the graph is extended with the nodes and edges corresponding to previously unseen validation patients. To prevent data leakage, validation nodes are allowed to receive messages from training nodes but are never allowed to propagate messages to other validation or training nodes. Consequently, the prediction for each validation patient remains independent of the other validation patients. The same procedure is applied during testing. This setup reflects a realistic deployment scenario in which new patients are integrated into an existing graph without affecting the representations of other patients, whether already in the graph or newly arriving alongside them.

\section{Experimental setup}
\label{sec:setup}
\subsection{Datasets and Tasks}
\subsubsection{MIMIC-IV}
We used the MIMIC-IV database (version 3.1)~\cite{PhysioNet-mimiciv-3.1}, a publicly available dataset containing de-identified records of patients admitted to Beth Israel Deaconess Medical Center (BIDMC) in Boston, MA, USA, between 2008 and 2022. We used both the hospital and ICU modules of MIMIC-IV, together with MIMIC-CXR-JPG for chest X-ray imaging and MIMIC-IV-Note for clinical text. We defined three prediction tasks from this dataset, two of which share the same set of extracted predictors. Chest X-ray images were encoded using \textit{DenseNet121-res224-chex}~\cite{cohen2022torchxrayvision}, and multiple images per patient were aggregated using a Gated Recurrent Unit (GRU)~\cite{chung2014empirical} trained jointly with the prediction model. Clinical notes were encoded using \textit{BioClinicalBERT}\footnote{\url{https://huggingface.co/emilyalsentzer/Bio_ClinicalBERT}}. Detailed cohort construction and variable selection are provided in~\ref{app:mimic}.

\paragraph{30-day readmission risk.}
Thirty-day hospital readmission is commonly used as an indicator of quality of care, as reducing avoidable readmissions can improve patient outcomes and lower healthcare costs. Following~\cite{tang2023predicting}, we formulated this task as a binary classification problem, where a patient is labeled positive if readmitted within 30 days of hospital discharge and negative otherwise. Patients who died within 30 days of discharge are also labeled positive. Each sample corresponds to a single eligible hospital admission. For patients with multiple eligible admissions, one admission was randomly selected, resulting in $N=31{,}288$ patients, with an outcome prevalence of $13\%$. The considered modalities are demographics (7 features), laboratory tests (46), diagnoses (269), chest X-rays (1,042-dimensional embeddings), medications (41), procedures (90), radiology notes (768-dimensional embeddings), and discharge notes (768-dimensional embeddings).

\paragraph{48-hour ICU length of stay.}
Accurate length-of-stay prediction can support hospital resource allocation and discharge planning. Following~\cite{soenksen2022integrated}, we formulated this task as binary classification, predicting whether a patient admitted to the intensive care unit (ICU) will be discharged alive within the following 48 hours. Patients who die during this window are assigned to the negative class. Each sample corresponds to a single ICU stay. For patients with multiple ICU stays, one stay was randomly selected, resulting in $N=13{,}235$ patients, with an outcome prevalence of $13\%$. The considered modalities are demographics (6 features), laboratory tests (22), chest X-rays (1,042-dimensional embeddings), radiology notes (768-dimensional embeddings), ICU procedures (10), and ICU chart events (9).

\paragraph{48-hour ICU mortality risk.}
ICU care requires rapid assessment of patient deterioration, making early mortality prediction a potentially valuable aid to clinical decision-making. Following~\cite{soenksen2022integrated}, we formulated this task as binary classification, predicting whether a patient admitted to the ICU will die within the following 48 hours. This task uses the same cohort as the 48-hour ICU length of stay task ($N=13{,}235$ patients), with an outcome prevalence of $3\%$, and the same modalities.

\subsubsection{NIH-TB}
The NIH-TB dataset~\cite{rosenthal2017tb}, provided through the National Institute of Allergy and Infectious Diseases (NIAID) Tuberculosis Portals Program, contains de-identified tuberculosis patient records collected across multiple countries. We used the preprocessed version described in~\cite{d2022fusing,d2023maxcorrmgnn,d2024fusing}, publicly available in their GitHub repository.\footnote{\url{https://github.com/IBM/MMMG/blob/main/TB_data.pickle.zip}} It contains 3,051 patients and five modalities: demographics (28-dimensional embeddings), clinical data (1,400), treatment regimen (233), genomic data (4,497), and chest CT imaging (2,048). The dataset also includes 8 miscellaneous continuous features. Genomic data include SNP features available for all patients and functional genomic features available for 254 patients, while CT scans are available for 1,015 patients. Further details on the preprocessing can be found in~\cite{d2022fusing,d2023maxcorrmgnn,d2024fusing}. Missing chest CT and functional genomic features were imputed using the column-wise mean. We used this dataset for two classification tasks.

\paragraph{Tuberculosis treatment outcome.}
Early identification of patients at risk of poor outcomes following tuberculosis treatment can help clinicians assess disease severity and support more targeted treatment and monitoring. This is a multiclass classification task with five possible treatment outcomes: Died, Still on Treatment, Completed, Cured, and Failure, occurring with proportions $21\%$, $9\%$, $50\%$, $11\%$, and $9\%$, respectively.

\paragraph{Mortality risk after tuberculosis treatment.}
This task is a binary special case of the previous task, in which we predict death versus all other treatment outcomes. The outcome prevalence is $21\%$.

\subsection{Baselines}
We compare DIGOM with three groups of multimodal fusion approaches to assess the contribution of graph-based interactions and the specific interaction types modeled by our framework. Specifically, DIGOM explicitly captures three types of interactions: (i) Intra-patient cross-modal interactions; (ii) Cross-patient same-modal interactions; and (iii) Cross-patient cross-modal interactions. Accordingly, we consider: (a) conventional fusion methods that do not explicitly model patient interactions, to assess the benefit of graph-based relational modeling; (b) graph-based methods that capture only a subset of these interaction types, to evaluate the contribution of the different interaction types; and (c) a graph-based method that models all three interaction types, to compare our graph construction and message-passing strategy with an existing method.

For DIGOM and all baselines, temporal modalities are first aggregated using a GRU trained jointly with the prediction model, except for STGNN, which incorporates temporal aggregation directly into its graph message-passing mechanism. Table~\ref{tab:interaction_types} summarizes the baselines considered and the interaction types modeled by each method.

\paragraph{Early Fusion.}
Individual modality features are concatenated and passed through a multilayer perceptron (MLP) to produce the final prediction.
 
\paragraph{Joint Fusion.}
Each modality is first projected independently using a dedicated MLP. The resulting modality-specific projections are then concatenated and passed through a second MLP to produce the final prediction. The entire model is trained end-to-end.
 
\paragraph{Late Fusion.}
Each modality is processed independently by its own MLP to produce a modality-specific prediction. The resulting predictions are then combined through a fully connected layer to obtain the final multimodal prediction.

\paragraph{Multimodal Spatiotemporal Graph Neural Network (STGNN).}
This baseline~\cite{tang2023predicting} models Cross-patient same-modal interactions. A patient population graph is constructed by computing similarities between demographic features and retaining the top-$k\%$ most similar patients prior to learning. The same graph connectivity is used across all modality views. For temporal modalities, STGNN incorporates graph-based spatial aggregation into the GRU by replacing its matrix multiplications with GNN layers. At each time step, message passing is performed over the graph using the modality information available at that time step, and the resulting spatial representations are aggregated across time using a GRU. For non-temporal modalities, message passing is performed once, at a single time step. The resulting modality-specific representations are fused across views using an MLP. As NIH-TB does not include temporal data, this baseline was evaluated only on the MIMIC-IV tasks. To ensure that the baseline graphs are at least as dense as the densest graphs allowed for DIGOM, $k$ was set to the highest value in the search range of $\rho$, the target density of DIGOM's density loss.
 
\paragraph{Multimodal Spatial Graph Neural Network (SGNN).}
SGNN is a non-temporal adaptation of STGNN that we constructed to evaluate the graph-based fusion principle underlying STGNN in settings without temporal data. The population graph is constructed using the same patient-similarity procedure, while temporal message passing is omitted. As in the other baselines, temporal feature aggregation, when applicable, is performed before message passing. For each modality, information is propagated across the corresponding patient graph, and the resulting modality-specific representations are subsequently fused using an MLP. This baseline therefore models Cross-patient same-modal interactions and can be applied to both temporal and non-temporal datasets.

\paragraph{Multimodal Graph Learning (MMGL).}
MMGL~\cite{zheng2022multi} first models Intra-patient cross-modal interactions using attention-based multimodal fusion, then constructs a population graph to capture cross-patient relationships. The fused patient representations are used to learn a single-relational patient graph via latent graph learning. Thus, MMGL captures Intra-patient cross-modal interactions and cross-patient interactions between patient-level representations, but does not distinguish cross-patient interactions by modality. The original method uses a learnable threshold and a sparsity-inducing loss to control graph density. However, on our MIMIC-IV cohorts, this mechanism produced graphs too dense to fit in memory during early training iterations. We therefore additionally applied a differentiable top-$k$ selection for MIMIC-IV, retaining the sparser of the two graphs produced by the differentiable top-$k$ and the original sparsity mechanism. The value of $k$ was set as for STGNN. For NIH-TB, graph density was not an issue, so the original sparsity mechanism was used.

\paragraph{Maximal Correlation Multi-layered Graph Neural Network (MaxCorrMGNN).}
MaxCorrMGNN~\cite{d2023maxcorrmgnn} jointly models all three interaction types through a multiplex graph. Intra-patient cross-modal and Cross-patient cross-modal interactions are represented by inter-planar connections, while Cross-patient same-modal interactions are represented by intra-planar connections. The graph structure and node representations are learned jointly, end-to-end, using maximal-correlation-based latent representations. As with MMGL, the original learnable edge threshold does not guarantee sufficiently sparse graphs, and the resulting graphs were too dense to fit in memory for our MIMIC-IV cohorts. We therefore applied the same differentiable top-$k$ strategy described above, retaining the sparser of the two resulting graphs. On NIH-TB, we report results for both the original method and its top-$k$ variant.

\begin{table}[t]
\centering
\setlength{\tabcolsep}{4pt}
\caption{Interaction types modeled by the considered baselines and DIGOM. $\checkmark$ indicates that the interaction type is explicitly modeled, $\sim$ indicates that it is captured but not modeled separately, and \ding{55} indicates that it is not modeled.}
\label{tab:interaction_types}
\begin{tabular}{lccc}
\toprule
\textbf{Method} 
& \textbf{Intra-patient} 
& \textbf{Cross-patient} 
& \textbf{Cross-patient} \\
& \textbf{cross-modal} 
& \textbf{same-modal} 
& \textbf{cross-modal} \\
\midrule
Early Fusion      & \ding{55} & \ding{55} & \ding{55} \\
Joint Fusion      & \ding{55} & \ding{55} & \ding{55} \\
Late Fusion       & \ding{55} & \ding{55} & \ding{55} \\
STGNN             & \ding{55} & \checkmark & \ding{55} \\
SGNN              & \ding{55} & \checkmark & \ding{55} \\
MMGL              & \checkmark & $\sim$ & $\sim$ \\
MaxCorrMGNN       & \checkmark & \checkmark & \checkmark \\
\midrule
DIGOM (ours)      & \checkmark & \checkmark & \checkmark \\
\bottomrule
\end{tabular}
\end{table}

\subsection{Evaluation protocol}
\begin{figure} 
\centering 
\includegraphics[width=\linewidth]{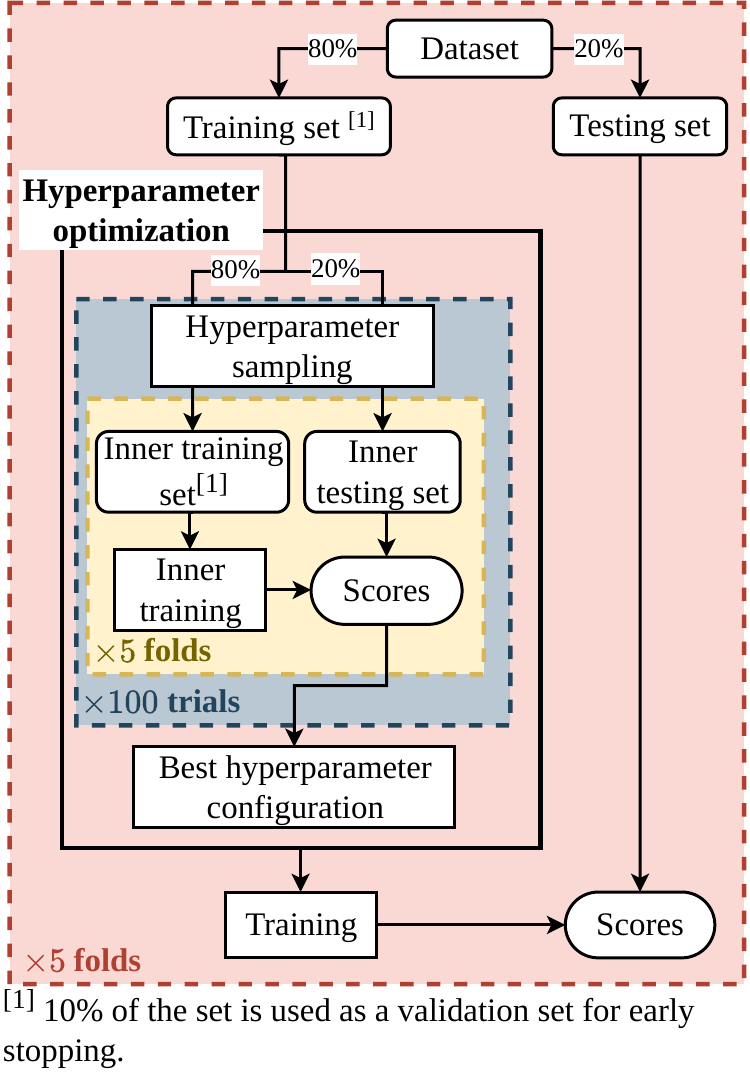}
\caption{Evaluation protocol of each model based on nested five-fold cross-validation. The inner loops are used for hyperparameter optimization, and the same data splits are used for all models.}
\label{fig:exp_setup}
\end{figure}

The proposed model and all baseline models were evaluated under an identical experimental protocol using a nested five-fold cross-validation, as shown in Figure~\ref{fig:exp_setup}. We divided the patients in the dataset using five-fold cross-validation into distinct training and testing sets. For MIMIC-IV, where patients may have multiple occurrences due to repeated hospital admissions or ICU stays, one occurrence per patient was randomly sampled and retained before performing cross-validation. Patients in each training set were further divided into distinct inner training and inner testing sets using an inner five-fold cross-validation. The inner sets were entirely dedicated to optimizing the hyperparameters of the models for each outer training fold. We created an additional validation set (as well as an inner validation set) for each of the five cross-validation splits, enabling us to track model performance throughout training and perform early stopping when the scores did not improve for a specified number of epochs. Each (inner) validation set was created by randomly sampling 10\% of patients from the corresponding (inner) training set.

For each model, hyperparameters were optimized by sampling 100 candidate configurations. Each configuration was evaluated by training the model on the five inner training sets and measuring the AUROC on the corresponding inner test sets; the configuration achieving the highest average inner AUROC was selected and used to retrain the model on the full outer training set. For the multiclass tuberculosis treatment outcome task, AUROC was computed as the macro-average of the one-vs-rest AUROCs over the five classes.

\subsubsection{Implementation details}
All models were optimized using the Adam optimizer with $\beta_1 = 0.9$, $\beta_2 = 0.999$, and a batch size of $1000$. Continuous features were standardized, and categorical features were one-hot encoded. A bidirectional GRU with a single layer was used to aggregate temporal modalities. For DIGOM, the projection functions $f_\phi^k$ and $f_\theta^k$ are implemented as three-layer MLPs, while $f_\psi$ and $g$ are linear transformations. The number of graph propagation layers $L$, the temperature $\tau$, the target density $\rho$, and the coefficient $\lambda_2$ were optimized during hyperparameter search, while $\lambda_1$ was fixed at $0.001$ and $\epsilon$ at $0$. The number of warm-up epochs $L_{\text{warmup}}$ was set to $15$ for 30-day readmission risk and $5$ for all other tasks. To optimize GPU memory usage, adjacency matrices were sampled in blocks of 8,000 rows, which were then concatenated. Since each entry is sampled independently, this block-wise procedure does not affect the resulting graph. Hyperparameter search ranges for DIGOM and all baselines are provided in~\ref{app:hps_values}. All experiments were performed using GPU acceleration, with DGL\footnote{\url{https://www.dgl.ai}} for graph construction and GNN modules, PyTorch\footnote{\url{https://pytorch.org}} for model training, and Optuna\footnote{\url{https://optuna.readthedocs.io}} with Ray\footnote{\url{https://www.ray.io}} for parallelized hyperparameter optimization.

\section{Results}
\label{sec:results}
\subsection{Performance comparisons}
\label{subsec:performance_comp}
We report in Table~\ref{tab:performance_comparison} the Area Under the Receiver Operating Characteristic Curve (AUROC) for DIGOM and the baseline models introduced above across the five prediction tasks. DIGOM achieved the highest AUROC on all five tasks, indicating that explicitly modeling multimodal interactions is generally beneficial, though the relative importance of each interaction type varies across prediction tasks. Overall, graph-based models tended to outperform conventional fusion methods, although this trend was not universal across all tasks and settings.

\begin{table*}[htbp]
\centering
\caption{AUROC comparison across the five prediction tasks. Results are reported as \textit{mean} $\pm$ \textit{standard deviation} across the five outer testing folds. For the multiclass tuberculosis treatment outcome, AUROC is macro-averaged over the five classes. The best-performing model is shown in \textbf{bold}, and the second-best is \underline{underlined}. Statistical significance was assessed using a one-sided Wilcoxon signed-rank test~\cite{wilcoxon1992individual} to determine whether each baseline significantly underperformed DIGOM. $^{*}$ and $^{\dagger}$ indicate $p\text{-value}<0.05$ and $0.05\leq p\text{-value}<0.1$, respectively. From left to right, the tasks are 48-hour ICU mortality risk, 48-hour ICU length of stay, 30-day readmission risk, tuberculosis treatment outcome, and mortality risk after tuberculosis treatment.}
\label{tab:performance_comparison}
\begin{tabular}{lccccc}
\toprule
 
& \multicolumn{3}{c}{\textbf{MIMIC-IV}} 
& \multicolumn{2}{c}{\textbf{NIH-TB}} \\
\cmidrule(lr){2-4} \cmidrule(lr){5-6}
 
& \textbf{48h Mortality}
& \textbf{48h LOS}
& \textbf{30d Readmission}
& \textbf{TB Outcome}
& \textbf{TB Mortality} \\
\midrule
 
Early Fusion
& $0.870 \pm 0.02^{*}$
& $0.802 \pm 0.02^{\dagger}$
& $0.675 \pm 0.01^{*}$
& $0.726 \pm 0.01^{*}$
& $0.734 \pm 0.01^{*}$ \\
 
Joint Fusion
& $0.871 \pm 0.01^{*}$
& $0.813 \pm 0.02$
& $0.670 \pm 0.02^{*}$
& $0.745 \pm 0.01^{\dagger}$
& $0.741 \pm 0.02^{*}$ \\
 
Late Fusion
& $0.865 \pm 0.01^{*}$
& $0.789 \pm 0.02^{*}$
& $0.672 \pm 0.01^{*}$
& $0.728 \pm 0.02^{*}$
& $0.745 \pm 0.02^{\dagger}$ \\
 
STGNN
& $0.887 \pm 0.01^{*}$
& $0.814 \pm 0.01$
& $0.675 \pm 0.01$
& --
& -- \\
 
SGNN
& $0.883 \pm 0.01^{*}$
& $0.813 \pm 0.01$
& $0.683 \pm 0.01$
& $0.735 \pm 0.01^{*}$
& \underline{$0.766 \pm 0.02$} \\
 
MMGL
& \underline{$0.897 \pm 0.01$}
& $0.813 \pm 0.02$
& $0.673 \pm 0.01^{\dagger}$
& $0.738 \pm 0.01^{*}$
& $0.746 \pm 0.02^{*}$ \\
 
MaxCorrMGNN
& --
& --
& --
& \underline{$0.758 \pm 0.01$}
& $0.764 \pm 0.02^{*}$ \\
 
MaxCorrMGNN top-$k$
& $0.887 \pm 0.01^{*}$
& \underline{$0.815 \pm 0.02$}
& \underline{$0.685 \pm 0.01$}
& $0.749 \pm 0.01^{\dagger}$
& $0.747 \pm 0.03^{*}$ \\
 
\midrule
 
\textbf{DIGOM (ours)}
& $\mathbf{0.901 \pm 0.01}$
& $\mathbf{0.816 \pm 0.02}$
& $\mathbf{0.686 \pm 0.01}$
& $\mathbf{0.761 \pm 0.01}$
& $\mathbf{0.771 \pm 0.02}$ \\
 
\bottomrule
\end{tabular}
\end{table*}
 
For 48-hour ICU mortality, DIGOM significantly outperformed the conventional fusion methods and the graph-based models that capture only Cross-patient same-modal interactions (STGNN, SGNN). It also outperformed the top-$k$ variant of MaxCorrMGNN, suggesting that the proposed patient-adaptive graph construction method can identify more informative connections than a fixed sparsity constraint. DIGOM achieved a slightly higher AUROC than MMGL, although the difference was not statistically significant. Since MMGL already models cross-patient relationships at the patient level, without separating them by modality, this suggests that further distinguishing Cross-patient same-modal from Cross-patient cross-modal interactions offers limited additional benefit for this task.
 
For 48-hour ICU length of stay, DIGOM achieved performance comparable to Joint Fusion. This suggests that explicitly modeling relational interactions provides limited additional predictive value for this task, with the multimodal patient representation alone capturing most of the information relevant to discharge prediction.
 
For 30-day readmission risk, DIGOM significantly outperformed all conventional fusion methods but did not significantly differ from STGNN or SGNN. This indicates that the Cross-patient same-modal relationships modeled by these two baselines capture most of the useful relational signal for this task, with limited additional benefit from explicitly modeling cross-modal interactions.
 
For tuberculosis treatment outcome, DIGOM achieved the highest AUROC. Its improvement over methods modeling fewer interaction types suggests that jointly capturing Intra-patient cross-modal, Cross-patient same-modal, and Cross-patient cross-modal relationships can be beneficial for distinguishing among the five treatment outcomes. Notably, DIGOM matched the performance of the original MaxCorrMGNN while using a substantially sparser graph. MaxCorrMGNN required over 60\% graph connectivity, whereas DIGOM achieved comparable results with only approximately 9\% connectivity (see~\ref{app:average_connectivity}). When MaxCorrMGNN was constrained to a similar sparsity level using top-$k$ selection, its AUROC dropped. This suggests that DIGOM's patient-adaptive graph learning is more effective at selecting informative relationships than a fixed sparsity constraint.
 
For tuberculosis mortality risk, DIGOM achieved the highest AUROC but was not statistically different from SGNN. This indicates that Cross-patient same-modal interactions capture most of the relational information useful for this binary outcome. DIGOM nevertheless significantly outperformed MMGL, MaxCorrMGNN, and its top-$k$ variant, all of which model additional interaction types. This suggests that additional relationships do not necessarily improve prediction and may be detrimental when the graph-learning mechanism does not effectively identify informative connections.
 
Overall, the results reveal a task-dependent role of relational structure. Richer interaction modeling provided the greatest benefit for 48-hour ICU mortality and tuberculosis treatment outcome, whereas Cross-patient same-modal interactions appeared sufficient for 30-day readmission and tuberculosis mortality. For 48-hour ICU length of stay, graph-based relational modeling provided little benefit over conventional multimodal fusion. Importantly, DIGOM adapted to these differences without requiring the relevant interaction types to be specified in advance, and consistently achieved the highest AUROC across all five tasks. Even where a given interaction type was not strictly necessary, modeling it did not appear to harm performance, suggesting that DIGOM's graph learning is able to select a set of connections that either improves performance or leaves it unaffected, rather than introducing noise.
 
Table~\ref{tab:nri_comparison} further reports the Net Reclassification Improvement (NRI), which measures whether DIGOM more appropriately reclassifies patients across risk categories compared with each baseline. Positive NRI values indicate improved overall reclassification, whereas negative values indicate worse reclassification. DIGOM yielded positive NRI values for nearly all baseline comparisons, except for SGNN on tuberculosis mortality risk and some baselines on 48-hour ICU length of stay, where the NRI was negative, consistent with the small non-significant AUROC differences observed for these comparisons above. 
 
\begin{table}[htbp]
\centering
\caption{Net Reclassification Improvement (NRI) of DIGOM over each baseline. NRI measures whether DIGOM more appropriately reclassifies patients across risk categories compared with each baseline. Positive NRI values indicate improved overall reclassification, whereas negative values indicate worse reclassification. Since NRI is defined for binary classification, the multiclass tuberculosis treatment outcome is not included in this analysis. The decision threshold is selected to maximize the difference between sensitivity and the false positive rate on the training set. \textit{48M}: 48-hour ICU mortality risk, \textit{48LOS}: 48-hour ICU length of stay, \textit{30DR}: 30-day readmission risk, \textit{TBM}: mortality risk after tuberculosis treatment.}
\label{tab:nri_comparison}
\small
\setlength{\tabcolsep}{3pt}
\begin{tabular}{lcccc}
\toprule
& \textbf{48M}
& \textbf{48LOS}
& \textbf{30DR}
& \textbf{TBM} \\
\midrule
 
Early Fusion  & $0.053$ & $0.014$ & $0.011$ & $0.025$ \\
Joint Fusion  & $0.057$ & $-0.005$ & $0.018$ & $0.043$ \\
Late Fusion   & $0.051$ & $0.035$ & $0.027$ & $0.007$ \\
STGNN         & $0.024$ & $-0.001$ & $0.008$ & -- \\
SGNN          & $0.018$ & $-0.009$ & $0.009$ & $-0.017$ \\
MMGL          & $0.304$ & $0.004$ & $0.036$ & $0.019$ \\
MaxCorrMGNN   & -- & -- & -- & $0.006$ \\
MaxCorrMGNN top-$k$ & $0.056$ & $0.001$ & $0.002$ & $0.020$ \\
 
\bottomrule
\end{tabular}
\end{table}

\subsection{Disentangling the contribution of each interaction type}
\begin{figure*} 
\centering 
\includegraphics[width=\linewidth]{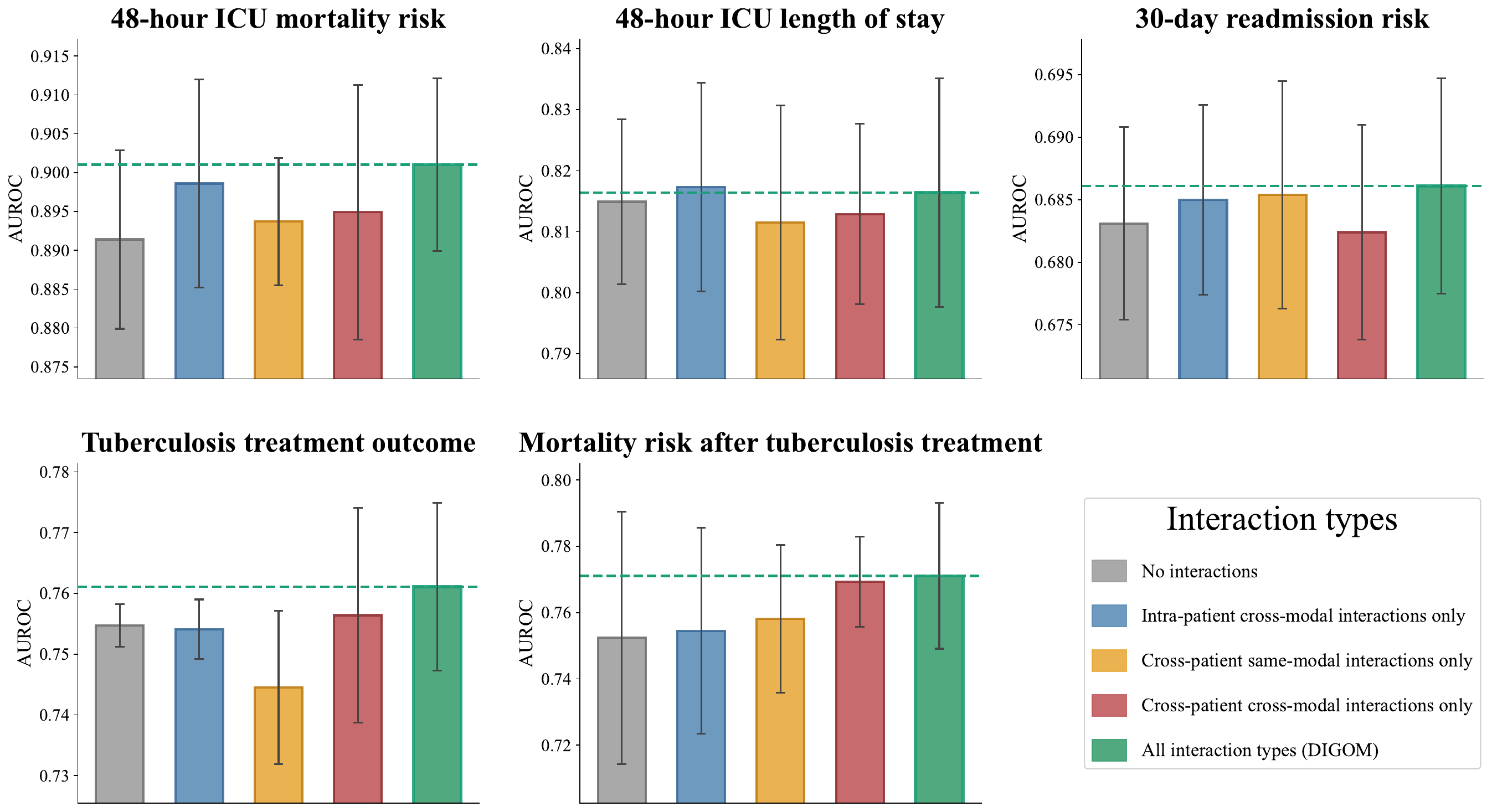}
\caption{AUROC of DIGOM trained with a single interaction type at a time, and with no interactions, across all five prediction tasks. Each bar reports the mean AUROC over five cross-validation folds, with error bars indicating the standard deviation. The dashed teal line marks the performance of the proposed DIGOM model, in which all three interaction types are learned jointly, and serves as a reference. The contribution of each interaction type varied across tasks, with no single type consistently providing the greatest benefit.}
\label{fig:interaction}
\end{figure*}

A key advantage of the disentangled formulation of DIGOM is that each interaction type can be independently evaluated by retaining only its corresponding graph views, without modifying the model architecture. We leveraged this property to assess the contribution of each interaction type by retraining DIGOM, with an unchanged architecture, using only one interaction type at a time, as well as with all connections removed. Figure~\ref{fig:interaction} reports the resulting AUROC for each variant relative to the proposed model, in which all three interaction types are learned jointly. The results are consistent with the performance comparisons in Table~\ref{tab:performance_comparison} and reveal substantial task-dependent differences in the use of relational information, with no single interaction type consistently providing the greatest benefit.
 
For 48-hour ICU mortality, using only Intra-patient cross-modal interactions resulted in the highest AUROC among the ablations, indicating that relationships between modalities within the same patient provide the most useful relational information for this task. However, both cross-patient interaction types also improved performance relative to the model without any interactions. This task therefore benefited from modeling all three interaction types together.
 
In contrast, relational information had a much smaller effect on 48-hour ICU length of stay, where using any single interaction type produced only a small change in AUROC compared with the model without interactions. This is consistent with the small performance differences between DIGOM and the baseline models in Table~\ref{tab:performance_comparison} and suggests that relational information provides limited additional value for this task. On the other hand, predicting 30-day readmission risk was primarily driven by Cross-patient same-modal interactions, followed by Intra-patient cross-modal interactions. This is consistent with the strong performance of SGNN and STGNN shown in Table~\ref{tab:performance_comparison}, which leverage Cross-patient same-modal interactions.
 
The two tuberculosis tasks showed a different pattern, both relying primarily on Cross-patient cross-modal interactions. This pattern was not observed in the MIMIC-IV tasks. One possible explanation is the more homogeneous nature of the NIH-TB cohort, where all patients share the same underlying disease. In this setting, information from other modalities of similar patients may provide useful context for predicting disease progression. This effect was strongest for tuberculosis mortality risk, where Cross-patient cross-modal interactions clearly outperformed the other interaction types, and was also observed, although to a lesser extent, for treatment outcome.
 
Interestingly, Cross-patient same-modal interactions contributed less than Cross-patient cross-modal interactions to tuberculosis mortality prediction in the DIGOM ablation, despite the strong performance of SGNN in Table~\ref{tab:performance_comparison}, which relies exclusively on this interaction type. This gap suggests that DIGOM may have under-leveraged this interaction type, or compensated for it with Cross-patient cross-modal interactions.
 
The tuberculosis treatment task further showed that the contributions of the three interaction types are not simply additive. When used individually, each interaction type provided only a limited improvement, or even reduced performance, relative to the model without connections. When combined, however, they produced a substantially larger gain. This suggests that the three interaction types capture complementary information and that their joint modeling is particularly beneficial for distinguishing among the five treatment outcomes.
 
Overall, these results confirm that the value of relational modeling depends on the task. Since no single interaction type was best across all tasks, they support learning the graph structure over all three interaction types jointly, rather than committing in advance to a fixed subset.

\subsection{Robustness to missing modalities}
\label{subsec:missing}
\begin{figure*} 
\centering 
\includegraphics[width=\linewidth]{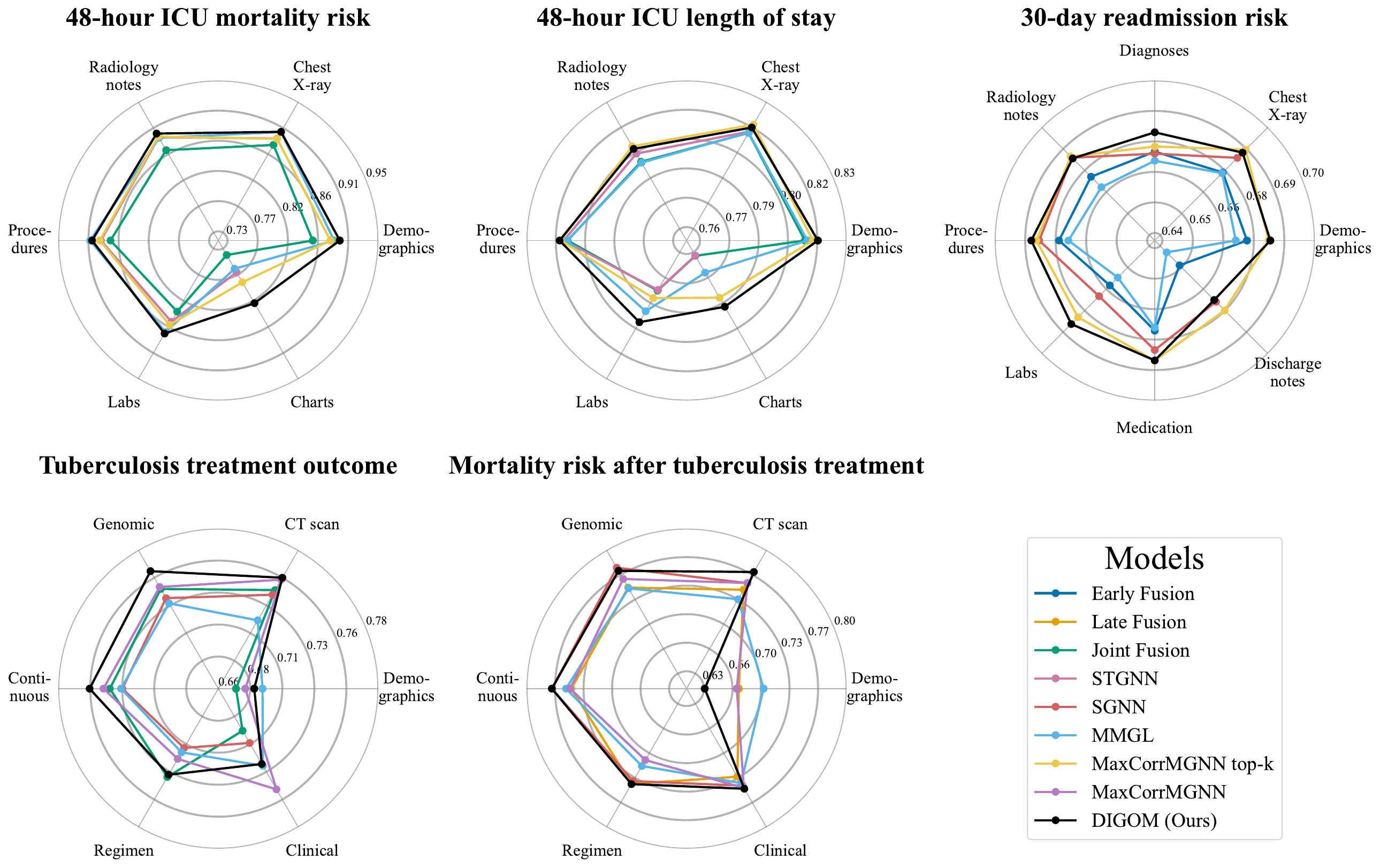}
\caption{AUROC of the selected baselines and DIGOM when one modality is masked at a time at inference time, across all five prediction tasks. For each masked modality, continuous columns are imputed using the column mean and categorical columns using the column median. Each spoke of the radar chart corresponds to a modality, and each line represents a model evaluated after masking that modality. Graph-based methods modeling multiple interaction types generally maintained higher performance under missing modalities than conventional fusion methods or methods restricted to a single interaction type.}
\label{fig:missing}
\end{figure*}

We next evaluated the robustness of the learned representations to missing modalities. This experiment simulated a deployment scenario in which a modality is unavailable for a given patient, and assessed whether the connections learned by our model can compensate for the missing information. For each group of baselines modeling the same interaction types, we selected the model with the highest AUROC in Table~\ref{tab:performance_comparison}. We then evaluated these models and DIGOM, all trained using the complete set of modalities, on the test set after masking one modality at a time. Figure~\ref{fig:missing} reports the resulting AUROC scores.
 
Overall, graph-based methods were more robust to missing modalities than conventional fusion methods. Methods that model multiple interaction types, such as MMGL, MaxCorrMGNN, and DIGOM, generally maintained higher performance under missing modalities than methods restricted to Cross-patient same-modal interactions, such as SGNN and STGNN. This suggests that modeling multiple types of connections provides additional sources of information that can compensate for missing modalities. Importantly, this benefit was not limited to tasks for which relational modeling improved performance with no missing data (Table~\ref{tab:performance_comparison}). For example, 48-hour ICU length of stay showed the smallest benefit from relational modeling when all modalities were available, yet Figure~\ref{fig:missing} shows that graph-based methods were substantially more robust to missing modalities than Joint Fusion. DIGOM was the most robust model for several masked modalities, particularly chart and laboratory events, which appear to be the most informative modalities for this task.
 
We also observed this consistent benefit of modeling different interaction types in the other MIMIC-IV tasks. For 48-hour ICU mortality, DIGOM and MMGL achieved similar performance when all modalities were available (Table~\ref{tab:performance_comparison}), but DIGOM was more robust to missing modalities, particularly chart events, which appear to be the most important modality for this task. This difference may be related to how cross-patient relationships are modeled in the two methods. MMGL connects patients without explicitly distinguishing the modalities involved in each interaction, whereas DIGOM directly connects modalities across patients. When a modality is missing, DIGOM can therefore propagate information from that same modality in related patients, providing a more direct source of information. In contrast, MMGL operates through higher-level patient representations. A similar pattern was observed for 30-day readmission, where DIGOM and MaxCorrMGNN showed comparable robustness across most missing-modality settings and generally outperformed SGNN, despite their similar performance with all modalities available. Together, these results suggest that the richer interaction types modeled by DIGOM provide useful alternative sources of information under missing data.
 
The tuberculosis tasks further support this observation, with one notable exception. Across most missing-modality settings, DIGOM achieved the highest AUROC for both tasks, confirming that its learned relationships can provide useful information when individual modalities are unavailable. When demographic data were missing, however, DIGOM showed the largest AUROC drop of any model on both tasks and performed worse than nearly all baselines, including conventional fusion methods. Demographics were by far the most informative modality for both tasks, which suggests that information from other modalities cannot fully compensate for the information lost when a highly informative modality is missing. We attribute this to the fact that, in DIGOM, each modality contributes both to a patient's own representation and to the graph used to propagate information across patients. Removing a modality therefore eliminates both sources of information at once, amplifying its impact relative to models in which modalities contribute more independently. 

\subsection{Impact of the density loss on graph sparsity}
\begin{figure*} 
\centering 
\includegraphics[width=\linewidth]{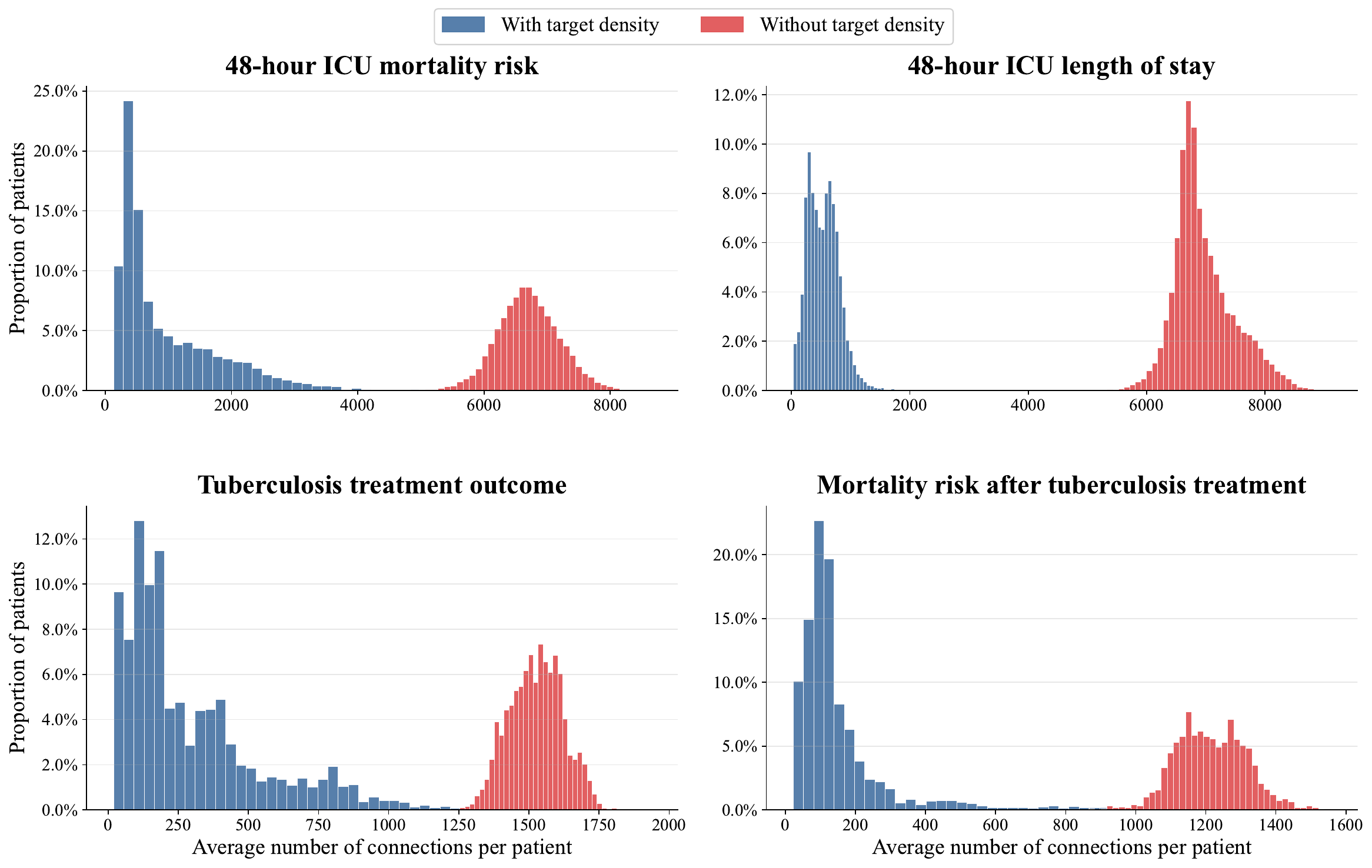}
\caption{Distribution of the average number of connections per patient in the graphs learned by DIGOM, with and without the density loss, across four prediction tasks. The distribution without the density loss could not be obtained for 30-day readmission risk due to the memory cost of constructing a dense graph for its large cohort; its distribution with the density loss is shown separately in Figure~\ref{fig:density_30dr}. The density loss produced substantially sparser graphs across all tasks, while Table~\ref{tab:ablation_density} shows that predictive performance remained largely unaffected, indicating that many connections can be pruned without loss of useful information.}
\label{fig:histogram}
\end{figure*}

Recall that we introduced the density loss in Equation~\ref{eq:density} to encourage sparse graphs and control their connectivity. We evaluated its impact on the sparsity of the learned graphs. Figure~\ref{fig:histogram} compares the distribution of the average number of connections per patient, obtained by averaging the number of connections of each modality node over all modalities of that patient, when DIGOM is trained with and without the density loss. A clear difference in connectivity was observed between the two settings, while overall AUROC remained similar (see~\ref{app:additional_results}). These results demonstrate that the density loss effectively reduced graph connectivity without affecting predictive performance. This further suggests that denser graphs are not necessarily required for accurate prediction, and that many connections can be removed without losing useful information, resulting in sparser graphs that require less memory and reduce the computational cost of training.
 
Importantly, the number of retained connections also varied across patients. Although the number of connections generally concentrated around the target density $\rho$, some patients had noticeably more and others fewer. This supports the patient-adaptive nature of DIGOM, as the model can select different numbers of neighbors for different patients rather than forcing a fixed connectivity threshold across the cohort.
 
\subsection{Post-hoc analysis of the learned interactions}
\label{subsec:posthoc}
\begin{figure*} 
\centering 
\includegraphics[width=\linewidth]{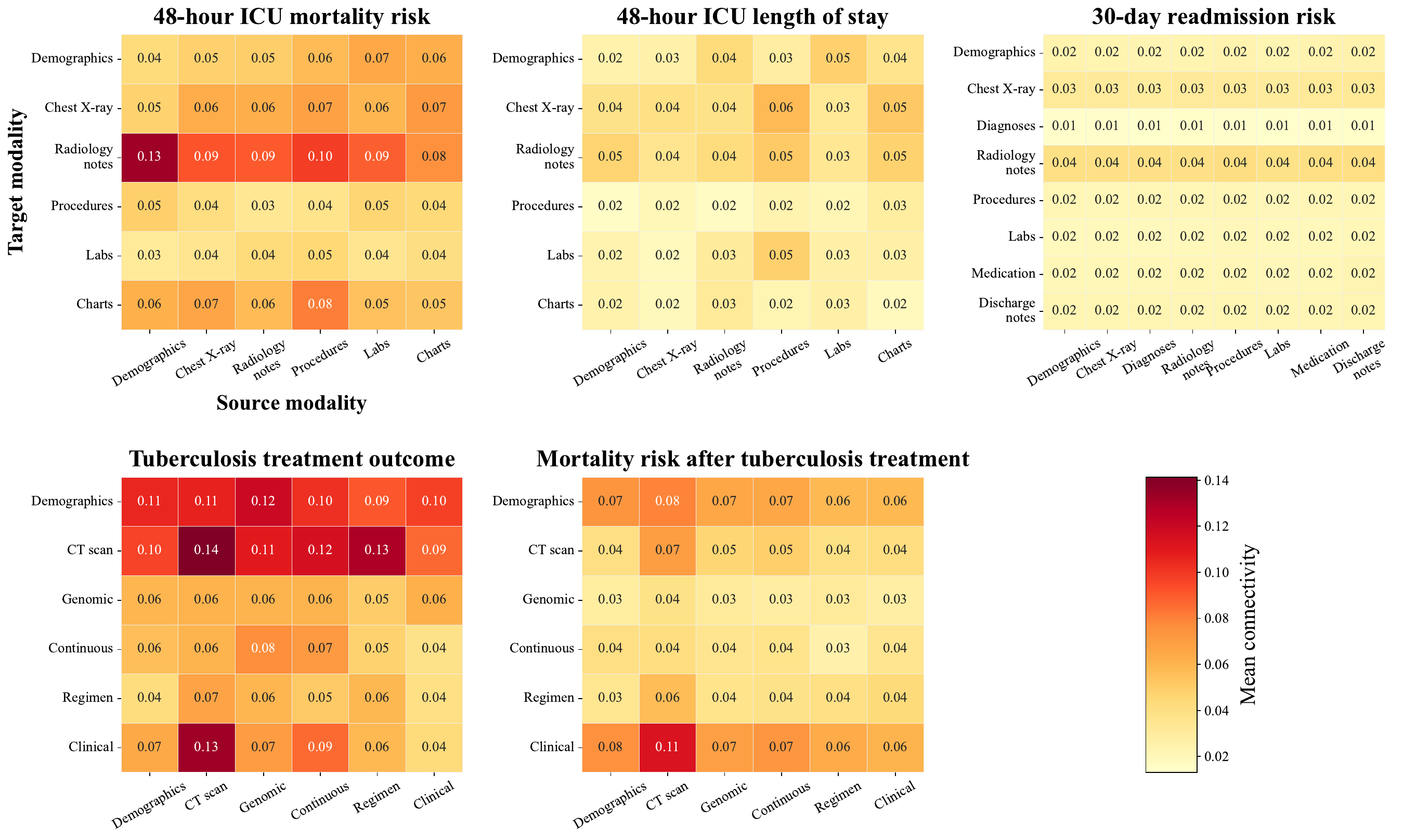}
\caption{Average connectivity between source and target modality pairs in the graphs learned by DIGOM across all five prediction tasks, averaged over the five test folds. Each cell reports the fraction of patients connected from the source modality to the target modality. Connectivity patterns varied substantially across tasks, with higher connectivity concentrated on CT scan, demographic, and clinical modalities for the tuberculosis tasks, and on radiology notes for 48-hour ICU mortality, reflecting task-specific differences in how the model retrieves relational information across patients and modalities.}
\label{fig:heatmap_con}
\end{figure*}

Finally, we analyzed the connectivity patterns learned between pairs of modalities to characterize how the model uses information across patients and modalities. Figure~\ref{fig:heatmap_con} reports the fraction of retained connections for each source-target modality pair. For example, a connectivity of $0.05$ from radiology notes to chest X-rays indicates that, on average, each target patient's chest X-ray representation aggregates information from approximately $5\%$ of source patients' radiology notes.
 
The learned connectivity patterns varied across tasks. For 48-hour ICU length of stay and 30-day readmission, connectivity was relatively evenly distributed across modality pairs, suggesting that the model did not consistently favor a small set of source-target relationships. In contrast, the tuberculosis tasks showed substantially higher connectivity for the CT scan, demographic, and clinical modalities, and radiology notes showed higher connectivity for 48-hour ICU mortality. This indicates that, for these modalities, the model retrieved information from a broader set of patients when making predictions.
 
Connectivity should not, however, be interpreted directly as the importance of a modality pair. Because incoming messages are averaged across neighbors, a larger number of connections spreads the propagated information across more sources and reduces the contribution of each one. High connectivity therefore reflects broader information retrieval, whereas low connectivity indicates that the model relies on a smaller and more targeted set of relationships.
 
\begin{figure*} 
\centering 
\includegraphics[width=\linewidth]{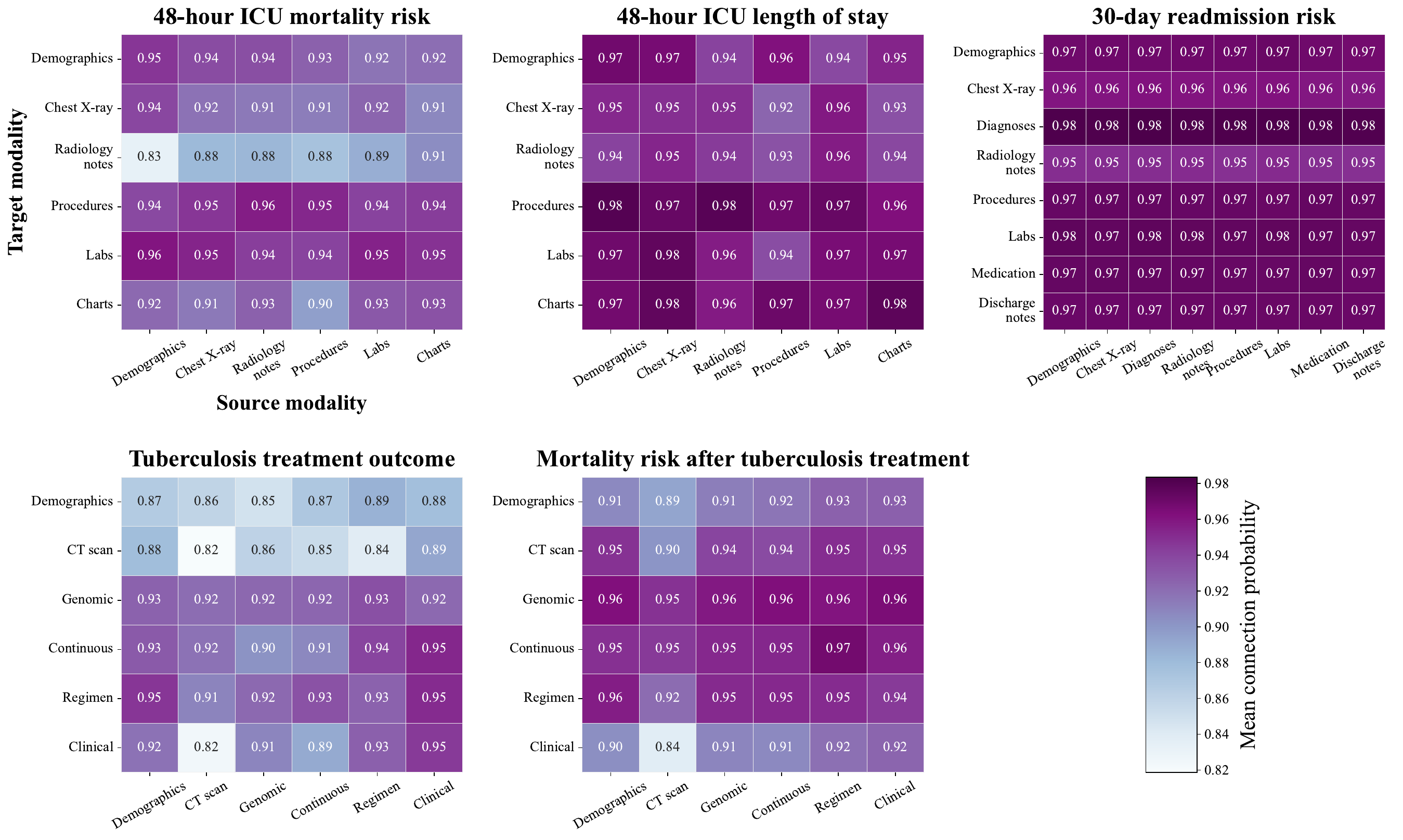}
\caption{Average connection probability between source and target modality pairs in the graphs learned by DIGOM across all five prediction tasks, averaged over the five test folds. Each cell reports the mean probability assigned to the edges selected between the source and target modality prior to sampling. Modality pairs with higher connectivity tended to exhibit lower connection probabilities, suggesting that the model selects connections with greater confidence when it can identify a small targeted set of informative relationships, and resorts to a broader but less certain set when the relevant connections are harder to identify.}
\label{fig:heatmap_proba}
\end{figure*}

We further examined the probabilities of the sampled edges during graph construction. Figure~\ref{fig:heatmap_proba} shows that higher connectivity was generally associated with lower connection probabilities, meaning that the model was less confident in its selected connections when it retained more of them. This suggests that when the model can identify the informative connections, it retains only a small, targeted set, and when it cannot, it retains a larger number of connections in an attempt to capture the relevant information among them.
 
We also examined connectivity separately for each outcome class in~\ref{app:posthoc}. The already elevated connectivity of demographics for tuberculosis mortality and of radiology notes for 48-hour ICU mortality was even higher in the positive class, along with chart events for ICU mortality. This suggests that the model found it comparatively harder to identify which patients contribute the most useful information for patients in the positive class. For tuberculosis treatment, the Died, Completed, and Cured classes showed the highest connectivity for CT scan and demographic modalities. These class-specific patterns further support the view that the learned graph structure should not be uniform across patients, tasks, or outcomes, but should adapt to each of them.

\section{Discussion}
\label{sec:discussion}
In this work, we propose DIGOM, a framework for modeling multimodal interactions in patient data by capturing relationships between modalities both within and across patients. Across five clinical prediction tasks and two healthcare datasets, DIGOM achieves the highest performance on every task. These results support our hypothesis that predictive performance depends not only on combining information from multiple modalities, but also on the relationships between them. Conventional fusion methods combine modality representations without modeling these relationships, while existing graph-based approaches typically capture only a subset of them. By jointly modeling intra-patient and cross-patient interactions within a single framework, DIGOM provides a broader representation of the clinical context around a patient.

The central empirical finding of this work is that no single interaction type consistently provides the largest contribution across tasks. Intra-patient cross-modal interactions are particularly beneficial for 48-hour ICU mortality, Cross-patient same-modal interactions for 30-day readmission, and Cross-patient cross-modal interactions for both tuberculosis tasks. For 48-hour ICU length of stay, individual interaction types provide limited additional benefit. This is also reflected in the baseline comparisons, where the second-best method varies across tasks. Methods relying on a predefined set of interactions, such as STGNN and SGNN for same-modal interactions, or MMGL for merged cross-patient interactions, therefore make an implicit assumption about which relational structure matters, an assumption that our results show does not hold consistently across clinical prediction tasks. This motivates learning the relevant interaction structure jointly with the prediction task, rather than fixing it a priori.

The comparisons with MMGL and MaxCorrMGNN further indicate that performance depends not only on which interaction types are modeled, but also on how connections are selected. MMGL, which models cross-patient relationships without distinguishing them by modality, underperforms specifically on tasks where SGNN, which relies only on Cross-patient same-modal interactions, performs strongly, such as 30-day readmission and tuberculosis mortality. This suggests that combining both interaction types may dilute the more informative one. On the other hand, MaxCorrMGNN models the same three interaction types as DIGOM, yet its top-$k$ variant, despite retaining a comparable number of edges (see~\ref{app:average_connectivity}), only matches DIGOM's performance on 30-day readmission and consistently underperforms on the other tasks where relational modeling is beneficial. Thus, at a comparable level of graph sparsity, DIGOM selects more informative connections than a fixed top-$k$ constraint. This suggests that top-$k$ selection, a standard approach for sparsifying graphs in the literature, is not well suited to identifying the most informative relationships.

The quality of selected connections, rather than their quantity, therefore appears to be the key factor. This is further illustrated by the tuberculosis tasks, where the non-sparsified MaxCorrMGNN requires more than 60\% of possible edges to reach performance comparable to DIGOM, whereas DIGOM achieves comparable or better results with substantially fewer edges (see~\ref{app:average_connectivity}). A denser graph increases the likelihood of retaining informative relationships, but also retains many uninformative ones. This interpretation is consistent with our density analysis, where removing the density loss produces substantially denser graphs while maintaining similar predictive performance. Together, these results indicate that many connections can be pruned without sacrificing accuracy, while reducing memory requirements and the computational cost of graph processing (see~\ref{app:inference_time}).

The missing-modality experiments provide further evidence for the value of relational modeling. Importantly, neither DIGOM nor the baselines are trained to handle missing modalities, so any robustness observed at inference time is a by-product of how they fuse modalities. Graph-based methods generally maintain higher performance than conventional fusion methods when modalities are masked at inference time. DIGOM is among the most robust approaches and generally outperforms methods modeling one or several interaction types across several missing-modality settings. This suggests that connections across patients and modalities can provide alternative contextual information when an observation is unavailable. Notably, this benefit is also observed for 48-hour ICU length of stay, where relational modeling provides limited improvement when all modalities are available but substantially improves robustness under missing modalities. Thus, relational modeling may provide a form of redundancy that can be useful when some modalities are missing for a patient. However, DIGOM is not equally robust to all missing modalities. For the tuberculosis cohort, removing the highly informative demographic modality substantially degrades DIGOM's performance, falling below the baseline methods for the task of mortality after tuberculosis treatment. Because each modality contributes both to a patient's own representation and to message passing, the model may learn to rely heavily on informative modalities in both. Retraining DIGOM without demographics yields better performance than removing this modality only at inference time (see~\ref{app:wo_demo}). This suggests that when a highly informative modality is unavailable, retraining the model without it may be preferable.

Another notable result concerns tuberculosis mortality, where SGNN, relying solely on Cross-patient same-modal interactions, nearly matches DIGOM, even though our ablation indicates a relatively limited contribution from this interaction type. This suggests that the interaction type that is most effective when modeled alone is not necessarily the one most strongly leveraged by the jointly optimized model. Jointly learning multiple interaction types may therefore lead to a highly informative overall solution without necessarily recovering the most parsimonious interaction structure for every task. Future work could investigate adaptive pruning to identify a task-specific set of relationships that is necessary and sufficient for prediction.

Beyond predictive performance, the disentangled structure of DIGOM naturally enables post-hoc analysis of the learned graph, as each interaction type is encoded in separate views and can be examined individually. Connectivity is concentrated on different modalities depending on the task, and some classes exhibit higher connectivity than others. These patterns provide insight into how the model distributes relational information and may support further investigation of the evidence used for prediction. However, the learned connections represent predictive associations rather than causal relationships and should not be interpreted as clinical explanations without further validation.

This work has several limitations worth addressing in future studies. First, modalities can interact for different reasons, including complementarity or redundancy, but DIGOM quantifies connectivity between two modalities solely based on the similarity of their latent representations. Future work could explore additional metrics better suited to capturing specific types of relationships, such as partial information decomposition to characterize redundancy and synergy. Second, although our density loss encourages sparse, patient-adaptive graphs rather than a fixed number of neighbors per node, our experiments show that the resulting graphs are not maximally pruned, as some retained connections may contribute little to prediction. Moreover, the target sparsity level is set prior to training rather than learned from the data. Future work should investigate training strategies that allow the model to determine the smallest set of connections sufficient for prediction without requiring a predefined sparsity level. In addition, all interaction types are currently weighted equally. Learning their relative importance jointly with the graph structure could further improve performance by allowing the model to emphasize the most informative interaction types for each task. Third, DIGOM is not trained to handle missing modalities, and its performance can degrade substantially when a highly informative modality is unavailable as shown in our previous experiments. A possible improvement would be to simulate missing modalities during training, allowing the model to learn representations that can better compensate for their absence and improve robustness in deployment settings. Finally, DIGOM models interactions between modalities and patients but does not capture relationships between individual features, whether within or across modalities. For example, it cannot model interactions between specific laboratory measurements, or between a particular procedure and a finding on a CT scan. Such an extension could provide a richer representation of clinical interactions, although it would substantially increase the number of candidate connections.

\printcredits

\section*{Data and code availability}
Software code allowing to run the experiments used to produce the results presented in this work is freely shared under the GNU General Public License v3.0 on the GitHub website at: \url{https://github.com/MEDomicsLab/DIGOM-study}. The MIMIC-IV database (\url{https://physionet.org/content/mimiciv}), MIMIC-CXR-JPG (\url{https://physionet.org/content/mimic-cxr-jpg}), and MIMIC-IV-Note (\url{https://physionet.org/content/mimic-iv-note}) are publicly available through PhysioNet upon completion of a data use agreement and the required training course. The NIH-TB dataset~\cite{rosenthal2017tb} is available through the NIAID Tuberculosis Portals Program (\url{https://tbportals.niaid.nih.gov}), and the preprocessed version used in this work is publicly available at \url{https://github.com/IBM/MMMG/blob/main/TB_data.pickle.zip}.

\section*{Funding sources}
This work was supported by the: (i) Natural Sciences and Engineering Research Council of Canada (NSERC), Discovery Grants program (\#RGPIN-2021-03996) and Alliance Research Projects program (\#ALLRP-608228-25); and (ii) Fonds de recherche du Québec (FRQ) - secteur Nature et technologies, NOVA – FRQ-NSERC program (https://doi.org/10.69777/378669). The funding sources had no role in the design and conduct of the study; collection, management, analysis, and interpretation of the data; preparation, review, or approval of the manuscript; or decision to submit the manuscript for publication.

\section*{Declaration of competing interests}
The authors declare no competing interests.

\appendix
\renewcommand{\thesection}{Appendix~\Alph{section}}
\renewcommand{\thetable}{\Alph{section}.\arabic{table}}
\renewcommand{\thefigure}{\Alph{section}.\arabic{figure}}
\renewcommand{\theequation}{\Alph{section}.\arabic{equation}}

\section{MIMIC-IV cohorts extraction}
\label{app:mimic}
\setcounter{table}{0}
\setcounter{figure}{0}
\setcounter{equation}{0}
This appendix details the cohort construction and variable selection procedures for the three MIMIC-IV prediction tasks.

\paragraph{30-day readmission risk.}
We identified eligible hospital admissions for which the patient underwent at least one frontal chest X-ray between arrival at the emergency department (ED) or hospital admission and discharge. Admissions were excluded if they were (a) ambulatory, emergency-unit, or direct observation stays, or (b) followed by a discharge to an acute care facility or another healthcare facility, or a discharge against medical advice. The demographic modality includes insurance, language, marital status, race, age, and gender. Missing demographic values were imputed with a new category ``unknown''. Length of stay is included as an additional variable within this modality. For laboratory measurements, we selected 46 laboratory tests following~\cite{tang2023predicting}. When multiple measurements were available during an admission, we retained the last measurement recorded before discharge. Missing measurements were imputed with mean values, and an additional binary presence indicator was included for each laboratory variable to distinguish observed measurements from missing values. This choice reflects the assumption that the absence of a laboratory measurement may itself be informative, for example, because the test was not considered clinically necessary. For medications, we selected 41 therapeutic classes following~\cite{tang2023predicting}. The resulting variables represent the number of occurrences of each medication class during the admission. For procedures and diagnoses, both ICD-9 and ICD-10 codes are present in MIMIC-IV due to the long observation period. To harmonize them, we converted ICD-9 codes to ICD-10 codes using General Equivalence Mappings (GEMs). To avoid excessive granularity, procedures were then grouped at level 2 of the ICD-10-PCS hierarchy, and diagnoses at level 2 of the ICD-10-CM hierarchy. Procedure and diagnosis variables are binary and indicate whether the corresponding procedure or diagnosis occurred during the admission.

\paragraph{48-hour mortality risk and length of stay in ICU.}
We selected all patients with an intensive care unit (ICU) stay and at least one frontal chest X-ray image. For both ICU prediction tasks, we used information available during the first 24 hours following ICU admission. This observation window was selected as a compromise between accumulating sufficient clinical information and enabling sufficiently early prediction of short-term outcomes. The demographic modality includes insurance, language, marital status, race, age, and gender. We selected 22 laboratory tests, 10 ICU-specific procedure events, and 9 ICU chart events following~\cite{soenksen2022integrated}. Procedure variables represent the number of occurrences of each event during the observation window, whereas laboratory and chart variables retain the last observed value. Missing laboratory and chart measurements were imputed with the mean value, together with a corresponding binary presence indicator.

\paragraph{Imaging and clinical text processing.}
Chest X-rays and clinical notes were processed using the same procedures across the prediction tasks. For chest X-rays, only frontal views were retained, and patients without a frontal chest X-ray were excluded from the corresponding cohort. Each retained image was embedded using the pretrained \textit{DenseNet121-res224-chex} model~\cite{cohen2022torchxrayvision}, yielding a 1,024-dimensional feature vector. Following~\cite{soenksen2022integrated}, the model prediction scores for 18 chest pathologies were concatenated with this vector, resulting in a 1,042-dimensional representation of each image. When multiple frontal images were available for an admission or ICU stay, we evaluated several strategies for aggregating the image representations (Table~\ref{tab:strategy_img}) and selected temporal aggregation, in which the sequence of image embeddings is processed in order of acquisition time by a GRU to obtain a single representation. For radiology and discharge notes, each note was processed using the pretrained \textit{BioClinicalBERT} model. Because \textit{BioClinicalBERT} accepts a maximum of 512 tokens, longer notes were divided into the minimum number of processable chunks, and the resulting chunk-level embeddings were averaged to obtain a single 768-dimensional representation per note. Each admission contains a single discharge note, whereas multiple radiology notes may be available. For radiology notes, we evaluated several aggregation strategies (Table~\ref{tab:strategy_notes}) and selected averaging of the individual note embeddings within each admission or ICU stay. For the $0.002\%$ of admissions in the 30-day readmission cohort and approximately $0.005\%$ of ICU stays without a radiology note, the embedding of an empty string was used.

\begin{table}
\centering
\caption{AUROC comparison across different aggregation strategies for chest X-ray imaging across the three prediction tasks in MIMIC-IV. The best strategy is shown in \textbf{bold}, and the second-best is \underline{underlined}. Results are reported as \textit{mean} $\pm$ \textit{standard deviation} across the five inner testing folds. \textit{Temporal} aggregates the image embeddings using a GRU, \textit{Average} computes their unweighted average, \textit{Weighted average} assigns higher weights to embeddings of images closer to the prediction time, and \textit{First} and \textit{Last} use only the first and last embeddings, respectively. \textit{48M}: 48-hour ICU mortality risk, \textit{48LOS}: 48-hour ICU length of stay, \textit{30DR}: 30-day readmission risk.}
\label{tab:strategy_img}
\small
\setlength{\tabcolsep}{3pt}
\begin{tabular}{lccc}
\toprule
& \textbf{48M} & \textbf{48LOS} & \textbf{30DR} \\
\midrule
Temporal        & $\mathbf{0.663 \pm 0.02}$ & \underline{$0.692 \pm 0.01$} & $\mathbf{0.611 \pm 0.01}$ \\
Average         & $\underline{0.661 \pm 0.02}$ & $0.680 \pm 0.03$ & $0.574 \pm 0.01$ \\
Weighted average & $0.625 \pm 0.03$ & $\mathbf{0.700 \pm 0.01}$ & $0.578 \pm 0.02$ \\
First           & $0.632 \pm 0.03$ & $0.629 \pm 0.02$ & $0.564 \pm 0.01$ \\
Last            & $0.639 \pm 0.03$ & $0.688 \pm 0.02$ & \underline{$0.576 \pm 0.01$} \\
\bottomrule
\end{tabular}
\end{table}
 
\begin{table}
\centering
\caption{AUROC comparison across different aggregation strategies for radiology notes across the three prediction tasks in MIMIC-IV. The best strategy is shown in \textbf{bold}, and the second-best is \underline{underlined}. Results are reported as \textit{mean} $\pm$ \textit{standard deviation} across the five inner testing folds. \textit{Temporal} aggregates the note embeddings using a GRU, \textit{Average} computes their unweighted average, \textit{Weighted average} assigns higher weights to embeddings of notes closer to the prediction time, and \textit{First} and \textit{Last} use only the first and last embeddings, respectively. \textit{48M}: 48-hour ICU mortality risk, \textit{48LOS}: 48-hour ICU length of stay, \textit{30DR}: 30-day readmission risk.}
\label{tab:strategy_notes}
\small
\setlength{\tabcolsep}{3pt}
\begin{tabular}{lccc}
\toprule
& \textbf{48M} & \textbf{48LOS} & \textbf{30DR} \\
\midrule
Temporal        & \underline{$0.716 \pm 0.02$} & \underline{$0.747 \pm 0.02$} & $\mathbf{0.641 \pm 0.01}$ \\
Average         & $\mathbf{0.718 \pm 0.01}$ & $\mathbf{0.751 \pm 0.01}$ & \underline{$0.634 \pm 0.01$} \\
Weighted average & $0.697 \pm 0.03$ & $0.736 \pm 0.02$ & $0.633 \pm 0.01$ \\
First           & $0.634 \pm 0.03$ & $0.629 \pm 0.01$ & $0.613 \pm 0.01$ \\
Last            & $0.677 \pm 0.01$ & $0.702 \pm 0.02$ & $0.603 \pm 0.02$ \\
\bottomrule
\end{tabular}
\end{table}

\section{Additional results}
\label{app:additional}
\setcounter{table}{0}
\setcounter{figure}{0}
\setcounter{equation}{0}
This appendix reports additional results complementing Section~\ref{sec:results}: graph connectivity for DIGOM and MaxCorrMGNN, the effect of the density loss, connectivity patterns by outcome class, inference time, and the performance of models retrained without demographics on the tuberculosis tasks.

\subsection{Graph connectivity of DIGOM and MaxCorrMGNN}
\label{app:average_connectivity}
Table~\ref{tab:connectivity} reports the average graph connectivity for DIGOM and MaxCorrMGNN across all prediction tasks under different sparsity constraints. For DIGOM, the target density acts as a soft regularization objective rather than a hard constraint, the actual connectivity may therefore differ from the target value across tasks and folds.

\begin{table*}[H]
\centering
\caption{Average connectivity rate for DIGOM and MaxCorrMGNN across prediction tasks, with and without graph density constraints. For MaxCorrMGNN, connectivity is reported as the maximum between the two propagation graphs $A \cdot C$ and $C \cdot A$, where $A$ captures Cross-patient same-modal similarities and $C$ captures cross-modal similarities. Note that the top-$k$ sparsity constraint is applied independently to $A$ and $C$. Thus, the products $A \cdot C$ and $C \cdot A$ yield a different effective sparsity, which is why the reported connectivity for MaxCorrMGNN with top-$k$ does not directly reflect $k$. ``--'' indicates that the condition could not be evaluated due to memory constraints from the large cohort size.}
\label{tab:connectivity}
\begin{tabular}{lccccc}
\toprule
& \textbf{48h Mortality} & \textbf{48h LOS} & \textbf{30d Readmission} & \textbf{TB Outcome} & \textbf{TB Mortality} \\
\midrule
DIGOM w/ density loss  & $7.42\%$  & $4.06\%$  & $3.01\%$ & $9.65\%$  & $4.99\%$  \\
DIGOM w/o density loss & $50.62\%$ & $52.63\%$ & --       & $50.03\%$ & $39.82\%$ \\
MaxCorrMGNN top-$k$      & $5.94\%$  & $6.33\%$  & $0.99\%$ & $5.01\%$  & $4.23\%$  \\
MaxCorrMGNN              & --        & --        & --       & $62.68\%$ & $63.48\%$ \\
\bottomrule
\end{tabular}
\end{table*}

\subsection{Effect of the density loss on connectivity and performance}
\label{app:additional_results}
\begin{figure} 
\centering 
\includegraphics[width=\linewidth]{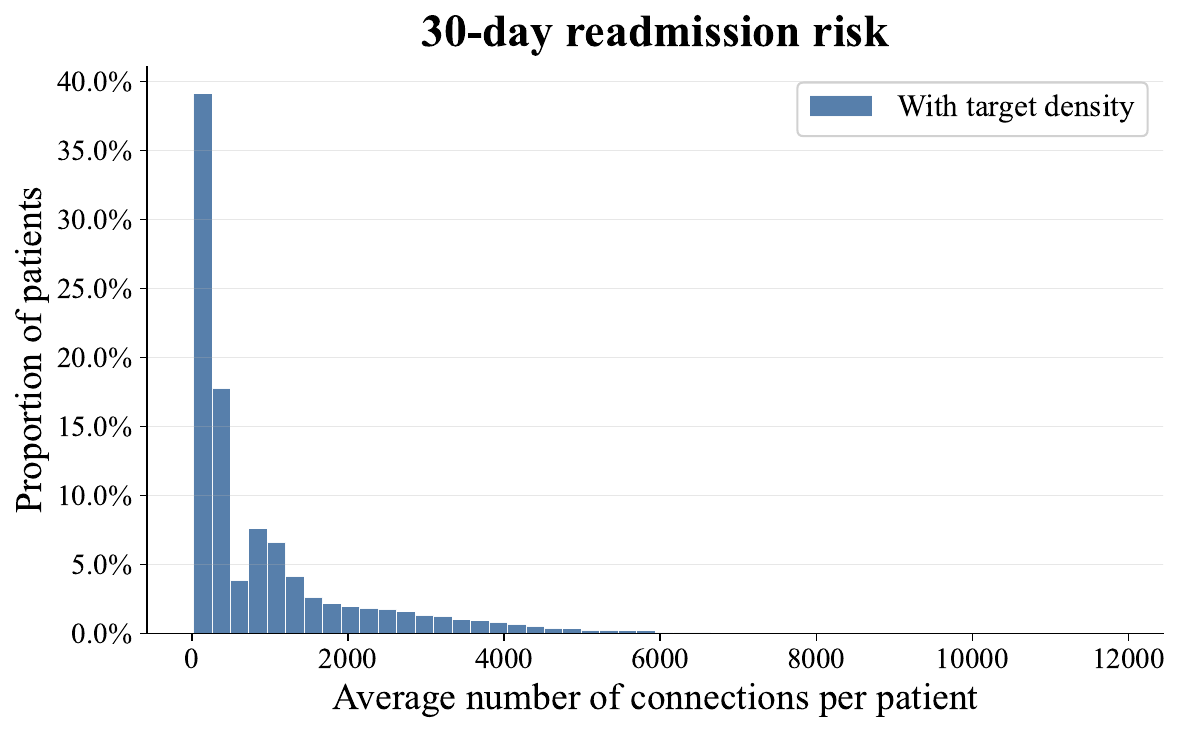}
\caption{Distribution of the average number of connections per patient in the graphs learned by DIGOM for 30-day readmission risk prediction. The density loss effectively constrains graph connectivity despite the large cohort size, producing a sparse graph concentrated around the target density $\rho$.}
\label{fig:density_30dr}
\end{figure}

This subsection investigates the effect of the density loss introduced in Equation~\ref{eq:density} on graph sparsity and predictive performance. Figure~\ref{fig:density_30dr} shows the distribution of the number of connections per patient in the graphs learned by DIGOM for 30-day readmission risk prediction. Unlike for the other tasks, we could not obtain this distribution without the density loss, as the resulting dense graph exceeded the available memory for this large cohort. Table~\ref{tab:ablation_density} further reports the effect of the density loss on AUROC across the remaining four tasks.

\begin{table}
\centering
\caption{Effect of the density loss in DIGOM on AUROC. Results are reported as \textit{mean} $\pm$ \textit{standard deviation} across the five outer testing folds. The 30-day readmission task could not be evaluated without the density loss due to the prohibitive memory cost of constructing a dense graph for its large cohort. \textit{48M}: 48-hour ICU mortality risk, \textit{48LOS}: 48-hour ICU length of stay, \textit{TBO}: tuberculosis treatment outcome, and \textit{TBM}: mortality risk after tuberculosis treatment.}
\label{tab:ablation_density}
\begin{tabular}{lcc}
\toprule
& \textbf{w/o density loss} & \textbf{with density loss} \\
\midrule
48M  & $0.895 \pm 0.01$ & $0.901 \pm 0.01$ \\
48LOS & $0.816 \pm 0.02$ & $0.816 \pm 0.02$ \\
TBO  & $0.752 \pm 0.01$ & $0.761 \pm 0.01$ \\
TBM  & $0.758 \pm 0.04$ & $0.771 \pm 0.02$ \\
\bottomrule
\end{tabular}
\end{table}

\subsection{Connectivity patterns by outcome class}
\label{app:posthoc}
\begin{figure*} 
\centering 
\includegraphics[width=\linewidth]{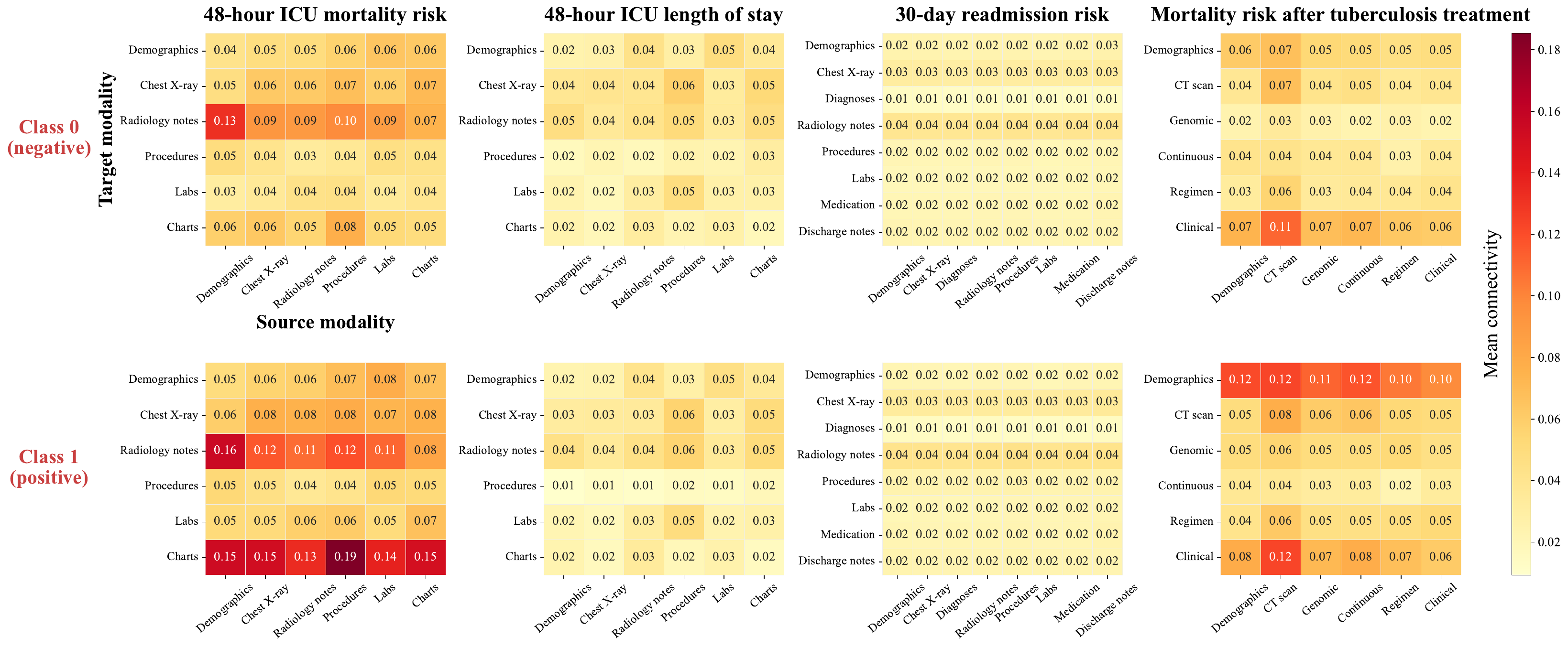}
\caption{Average connectivity between source and target modality pairs in the graphs learned by DIGOM, reported separately for the negative and positive classes across the four binary prediction tasks, averaged over the five test folds. For 48-hour ICU mortality and tuberculosis mortality risk, connectivity is higher in the positive class for the most connected modality pairs, suggesting that the model retrieves information from a broader set of patients when predicting the more clinically severe outcome.}
\label{fig:heatmap_binary}
\end{figure*}

\begin{figure}
\centering
\includegraphics[height=0.85\textheight, keepaspectratio]{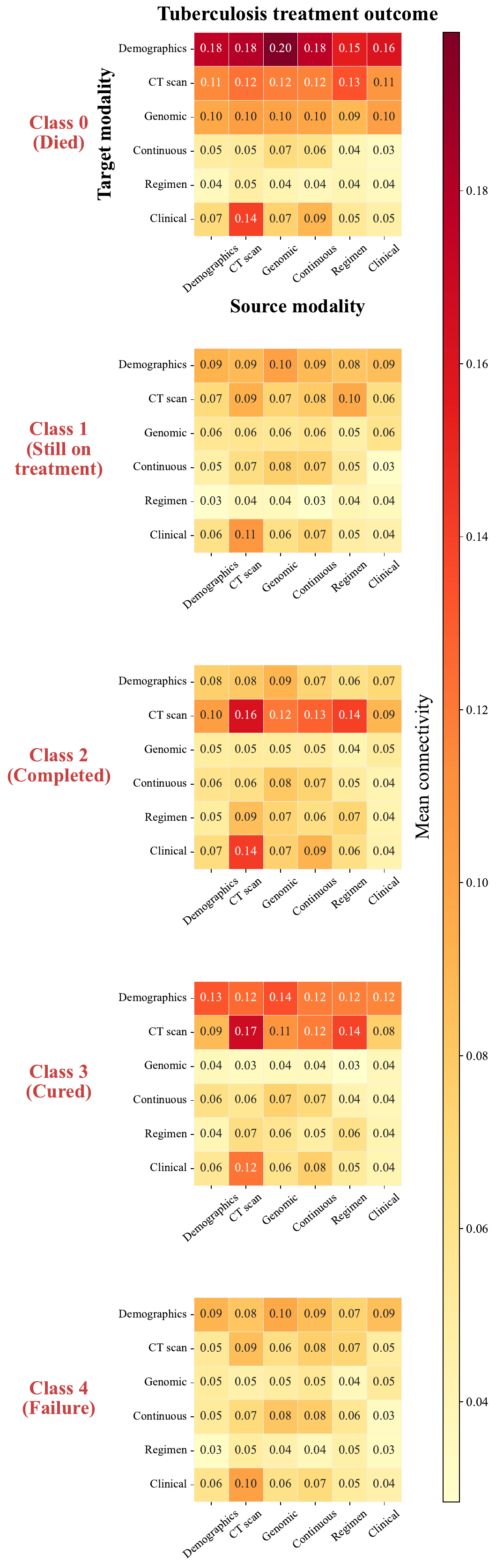}
\caption{Average connectivity between source and target modality pairs in the graphs learned by DIGOM, reported separately for each of the five tuberculosis treatment outcome classes, averaged over the five test folds. Connectivity is highest for the Died, Completed, and Cured classes, particularly for CT scan and demographic modalities, suggesting that the model retrieves information from a broader set of patients when predicting these outcomes.}
\label{fig:heatmap_tub}
\end{figure}

This subsection extends the post-hoc connectivity analysis of Section~\ref{subsec:posthoc} by reporting connectivity patterns separately for each outcome class. Figures~\ref{fig:heatmap_binary} and~\ref{fig:heatmap_tub} show the average connectivity between source and target modality pairs for the binary tasks and the tuberculosis treatment outcome task, respectively.
 
\subsection{Inference time of DIGOM and MaxCorrMGNN}
\label{app:inference_time}
Table~\ref{tab:inference_time} reports the average inference time on the test set for DIGOM and MaxCorrMGNN across all prediction tasks, with and without graph density constraints. The density loss reduces the inference time of DIGOM by a factor of two to four compared with training without it, consistent with the substantially sparser graphs reported in Table~\ref{tab:connectivity}. DIGOM is also faster than the original MaxCorrMGNN on the tuberculosis tasks. However, DIGOM remains slower than the top-$k$ variant of MaxCorrMGNN on most tasks, particularly for 30-day readmission, which is consistent with the higher connectivity of DIGOM's graphs on these tasks.

\begin{table*}[htbp]
\centering
\caption{Average inference time (in seconds) on the test set for DIGOM and MaxCorrMGNN across prediction tasks, with and without graph density constraints. Times are averaged across the five outer test folds. ``--'' indicates that the condition could not be evaluated due to memory constraints from the large cohort size.}
\label{tab:inference_time}
\begin{tabular}{lccccc}
\toprule
\textbf{Model} & \textbf{48h Mortality} & \textbf{48h LOS} & \textbf{30d Readmission} & \textbf{TB Outcome} & \textbf{TB Mortality} \\
\midrule
DIGOM w/ density loss  & $13.77$ & $11.55$ & $105.97$ & $0.83$ & $0.78$ \\
DIGOM w/o density loss & $44.32$ & $45.91$ & --       & $1.91$ & $1.98$ \\
MaxCorrMGNN top-$k$      & $11.46$ & $12.13$ & $32.78$  & $0.65$ & $0.56$ \\
MaxCorrMGNN              & --      & --      & --       & $5.89$ & $6.01$ \\
\bottomrule
\end{tabular}
\end{table*}

\subsection{Performance on tuberculosis tasks when retraining without demographics}
\label{app:wo_demo}
Table~\ref{tab:ablation_demographics} reports the performance of DIGOM and the selected baseline models on the tuberculosis tasks when the demographic modality is excluded during training, rather than masked only at test time. As discussed in Section~\ref{subsec:missing}, demographic information is by far the most informative modality for both tuberculosis tasks, and masking it at inference time leads to a substantial performance drop for DIGOM. Compared with masking demographics at inference time only, retraining without them improves performance for all models, particularly for DIGOM.

\begin{table}
\centering
\caption{AUROC on the tuberculosis tasks when models are retrained without the demographic modality. Only the models included in the missing-modality analysis are reported; conventional fusion is therefore represented by its best-performing variant for each task. Results are reported as \textit{mean} $\pm$ \textit{standard deviation} across the five outer testing folds.}
\label{tab:ablation_demographics}
\begin{tabular}{lcc}
\toprule
\textbf{Model} & \textbf{TB Outcome} & \textbf{TB Mortality} \\
\midrule
Joint Fusion          & $0.702 \pm 0.01$ & -- \\
Late Fusion           & -- & $0.723 \pm 0.02$ \\
SGNN                  & $0.721 \pm 0.01$ & $0.734 \pm 0.02$ \\
MMGL                  & $0.705 \pm 0.01$ & $0.718 \pm 0.02$ \\
MaxCorrMGNN           & $0.723 \pm 0.01$ & $0.707 \pm 0.02$ \\
\midrule
\textbf{DIGOM (ours)} & $0.728 \pm 0.01$ & $0.727 \pm 0.02$ \\
\bottomrule
\end{tabular}
\end{table}

\section{Hyperparameter values}
\label{app:hps_values}
\setcounter{table}{0}
\setcounter{figure}{0}
\setcounter{equation}{0}
We provide the hyperparameter search spaces used for DIGOM (Table~\ref{tab:hp_ranges_DIGOM}) and each baseline model (Table~\ref{tab:hp_ranges}). Ranges are shared across tasks unless otherwise noted.

\begin{table}
\centering
\caption{Hyperparameter search ranges used by DIGOM. Ranges are shared across tasks unless otherwise noted.}
\label{tab:hp_ranges_DIGOM}
\begin{tabular}{ll}
\toprule
\textbf{Hyperparameter} & \textbf{Values} \\
\midrule
Learning rate & $[0.0001, 0.01]$ \\
Positive class weight & $[0.1, 0.9]$ \\
GRU hidden size & $\{64, 128, 256, 512, 1024\}$ \\
$\rho$ (target density) & $\{0.01, 0.02, 0.03, 0.04, 0.05\}$$^{*}$ \\
$\lambda_1$ & $0.001$ \\
$\lambda_2$ & $[0.01, 0.5]$ \\
$f_{\theta}$ hidden size & $\{64, 128, 256\}$ \\
$f_{\phi}$ hidden size & $\{64, 128, 256\}$ \\
$f_{\psi}$ hidden size & $\{64, 128, 256\}$ \\
GNN layers $L$ & $[1, 3]$ \\
$\tau$ & $\{0.01, 0.1, 0.2\}$ \\
$\tau_{gs}$ (Gumbel-Softmax) & $\{0.01, 0.1, 1.0\}$ \\
\bottomrule
\end{tabular}

\vspace{2pt}
\footnotesize $^{*}$For 30-day readmission risk, $\rho \in \{0.001, 0.002, 0.003, 0.004, 0.005\}$ due to the larger cohort size.
\end{table}

\begin{table*}[H]
\centering
\caption{Hyperparameter search ranges used for each baseline model. Ranges are shared across tasks unless otherwise noted.}
\label{tab:hp_ranges}
\begin{tabular}{lll}
\toprule
\textbf{Model} & \textbf{Hyperparameter} & \textbf{Values} \\
\midrule

Common to all models
& Learning rate & $[0.0001, 0.01]$ \\
& Positive class weight$^\dagger$ & $[0.1, 0.9]$ \\

\midrule

\multirow{4}{*}{Early Fusion}
& Dropout & $[0, 0.2]$ \\
& GRU hidden size & $\{64, 128, 256, 512, 1024\}$ \\
& MLP layers & $[1, 4]$ \\
& MLP hidden size & $\{64, 128, 256, 512, 1024\}$ \\

\midrule
\multirow{4}{*}{Late Fusion}
& Dropout & $[0, 0.2]$ \\
& GRU hidden size & $\{64, 128, 256, 512, 1024\}$ \\
& MLP layers & $[1, 4]$ \\
& MLP hidden size & $\{64, 128, 256, 512, 1024\}$ \\

\midrule
\multirow{6}{*}{Joint Fusion}
& Dropout & $[0, 0.2]$ \\
& GRU hidden size & $\{64, 128, 256, 512, 1024\}$ \\
& MLP fuser layers & $[1, 4]$ \\
& MLP fuser hidden size & $\{64, 128, 256, 512, 1024\}$ \\
& MLP unimodal layers & $[1, 4]$ \\
& MLP unimodal hidden size & $\{64, 128, 256, 512, 1024\}$ \\

\midrule
\multirow{7}{*}{STGNN}
& Dropout & $[0, 0.2]$ \\
& GRU hidden size & $\{16, 32, 64, 128\}$ \\
& MLP fuser layers & $[1, 3]$ \\
& MLP fuser hidden size & $\{64, 128, 256, 512\}$ \\
& GNN layers & $[1, 2]$ \\
& GNN hidden size & $\{64, 128, 256, 512\}$ \\
& top-$k\%$ & $0.05^{*}$ \\

\midrule
\multirow{7}{*}{SGNN}
& Dropout & $[0, 0.2]$ \\
& GRU hidden size & $\{64, 128, 256, 512, 1024\}$ \\
& MLP fuser layers & $[1, 3]$ \\
& MLP fuser hidden size & $\{64, 128, 256, 512\}$ \\
& GNN layers & $[1, 2]$ \\
& GNN hidden size & $\{64, 128, 256, 512\}$ \\
& top-$k\%$ & $0.05^{*}$ \\

\midrule
\multirow{10}{*}{MMGL}
& Dropout & $[0, 0.2]$ \\
& GNN hidden se & $[1, 3]$ \\
& GRU hidden size & $\{64, 128, 256, 512, 1024\}$ \\
& Attention heads & $[1, 10]$ \\
& Transformer hidden size & $\{16, 32, 64, 128\}$ \\
& Transformer layers & $[0, 1]$ \\
& $\theta_{\text{smooth}}$ & $[0, 1]$ \\
& $\theta_{\text{degree}}$ & $[0, 0.5]$ \\
& $\theta_{\text{sparsity}}$ & $[0, 0.001]$ \\
& top-$k\%$ & $0.05^{*}$ \\

\midrule
\multirow{8}{*}{MaxCorrMGNN}
& GRU hidden size & $\{64, 128, 256, 512, 1024\}$ \\
& HGR learning rate & $[0.0001, 0.01]$ \\
& Weight decay & $[0.0001, 0.01]$ \\
& HGR weight decay & $[0.0001, 0.01]$ \\
& $\lambda_1$ & $[0.01, 0.5]$ \\
& GNN hidden size & $\{64, 128, 256\}$ \\
& MLP projection size & $\{128, 256, 512\}$ \\
& top-$k\%$ & $0.05^{*}$ \\

\bottomrule
\end{tabular}

\vspace{2pt}
\footnotesize $^{\dagger}$Used with binary classification tasks only.

\footnotesize $^{*}$For 30-day readmission risk, top-$k\% = 0.005$, due to the larger cohort size.
\end{table*}

\clearpage
\section*{Declaration of generative AI use}
During the preparation of this work the authors used Claude to improve the readability and language of the manuscript. After using this tool/service, the author(s) reviewed and edited the content as needed and take(s) full responsibility for the content of the published article.

\bibliographystyle{unsrturl}
\bibliography{cas-refs}

\end{document}